\documentclass[journal]{IEEEtran}
\usepackage{array}
\usepackage{cite}
\usepackage{caption}
\usepackage{amsmath,graphicx}
\usepackage{epstopdf}
\usepackage{xcolor}
\usepackage{amssymb}
\usepackage{pifont}
\usepackage{booktabs}
\usepackage{array}
\usepackage{tabu}
\usepackage{booktabs}
\usepackage{graphicx}
\usepackage{subfigure}
\usepackage{bm}
\usepackage{multirow}
\usepackage{graphicx}
\usepackage{color}
\usepackage{float}
\usepackage{stfloats}
\usepackage[misc]{ifsym}
\usepackage{stmaryrd}
\usepackage{grffile}
\usepackage{makecell}
\usepackage{threeparttable}
\usepackage{bm}
\usepackage{enumitem}
\usepackage{enumerate}
\usepackage[pagebackref=true,breaklinks=true,letterpaper=true,colorlinks,bookmarks=false]{hyperref}

\begin{document}

\title{Dense Nested Attention Network for Infrared Small Target Detection}

\author{{Boyang Li, Chao Xiao, Longguang Wang, Yingqian Wang, Zaiping Lin, Miao Li, Wei An, Yulan Guo}

\thanks{This work was partially supported in part by the National Natural Science Foundation of China (Nos. 61972435, 61401474, 61921001, 62001478)}

\thanks{Boyang Li, Chao Xiao, Longguang Wang, Yingqian Wang, Zaiping Lin, Miao Li, Wei An, Yulan Guo are with the College of Electronic Science and Technology, National University of Defense Technology (NUDT), P. R. China. Emails: \{liboyang20, xiaochao12, wanglongguang15, wangyingqian16, linzaiping, lm8866, anwei, yulan.guo\}@nudt.edu.cn. (Corresponding author: Zaiping Lin, Miao Li)}
}

\markboth{Journal of \LaTeX\ Class Files,~Vol.~14, No.~8, August~2015}%
{Shell \MakeLowercase{\textit{et al.}}: Bare Demo of IEEEtran.cls for IEEE Journals}

\maketitle

\begin{abstract}
   Single-frame infrared small target (SIRST) detection aims at separating small targets from clutter backgrounds. With the advances of deep learning, CNN-based methods have yielded promising results in generic object detection due to their powerful modeling capability.
   However, existing CNN-based methods cannot be directly applied to infrared small targets since pooling layers in their networks could lead to the loss of targets in deep layers. To handle this problem, we propose a dense nested attention network (DNA-Net) in this paper. Specifically, we design a dense nested interactive module (DNIM) to achieve progressive interaction among high-level and low-level features. With the repetitive interaction in DNIM, the information of infrared small targets in deep layers can be maintained. Based on DNIM, we further propose a cascaded channel and spatial attention module (CSAM) to adaptively enhance multi-level features. With our DNA-Net, contextual information of small targets can be well incorporated and fully exploited by repetitive fusion and enhancement. Moreover, we develop an infrared small target dataset (namely, NUDT-SIRST) and propose a set of evaluation metrics to conduct comprehensive performance evaluation. Experiments on both public and our self-developed datasets demonstrate the effectiveness of our method. Compared to other state-of-the-art methods, our method achieves better performance in terms of probability of detection (${P}_{d}$), false-alarm rate (${F}_{a}$), and intersection of union ($IoU$).

\end{abstract}


\section{Introduction}\label{introduction}
\IEEEPARstart{S}{ingle}-frame infrared small target (SIRST) detection is widely used in many applications such as maritime surveillance \cite{1-Maritime-Surveillance,yingxinyi}, early warning systems \cite{2-early-warning,Matianlei}, and precise guidance \cite{3-anti-miss}.
Compared to generic object detection, infrared small target detection has several unique characteristics:
1) \textbf{Small:} Due to the long imaging distance, infrared targets are generally small, ranging from one pixel to tens of pixels in the images. 2) \textbf{Dim:} Infrared targets usually have low signal-to-clutter ratio (SCR) and are easily immersed in heavy noise and clutter background. 3) \textbf{Shapeless:} Infrared small targets have limited shape characteristics. 4) \textbf{Changeable:} The sizes and shapes of infrared targets vary a lot among different scenarios.

\begin{figure}
\centering
\includegraphics[width=8.8cm]{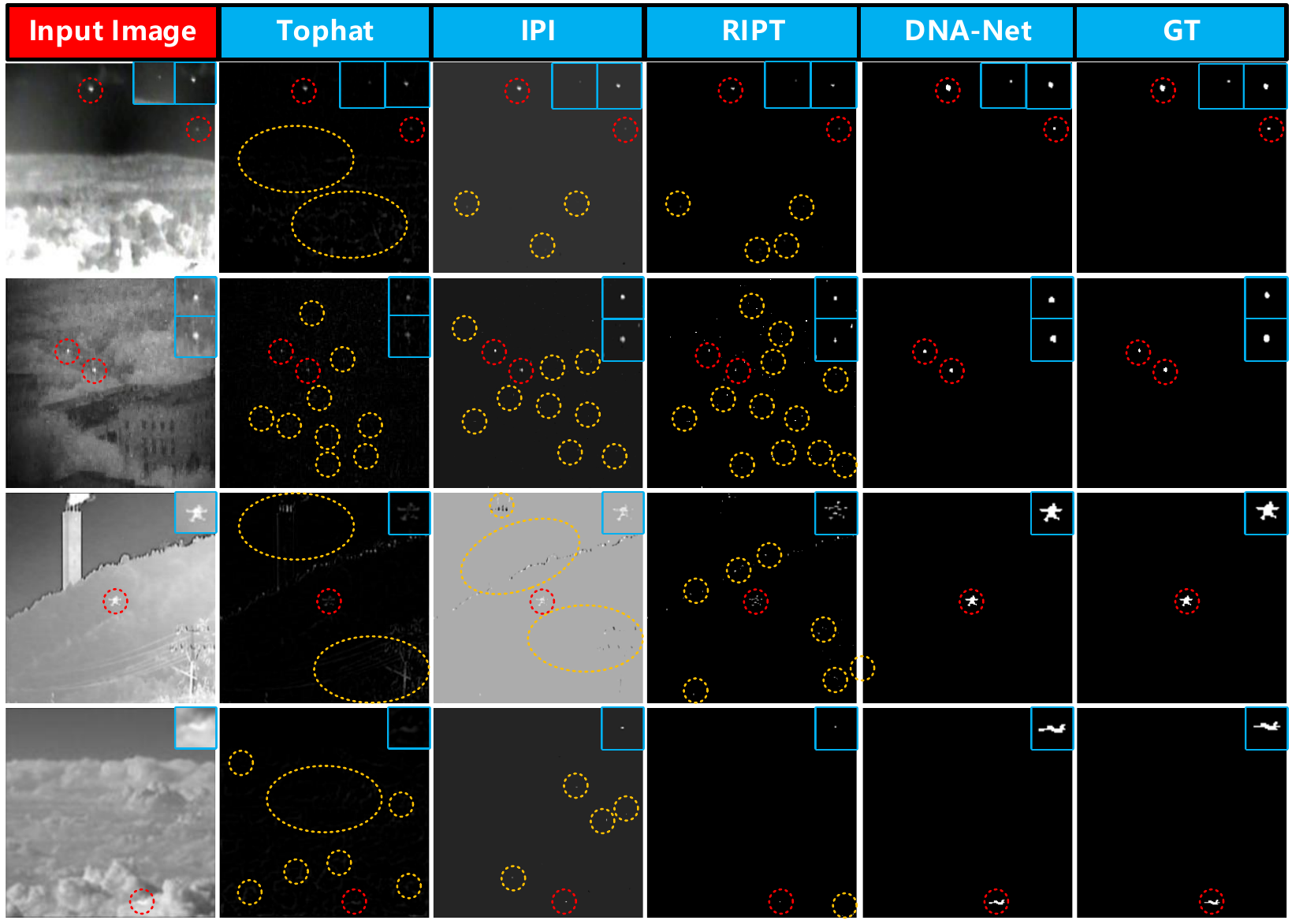}
\caption{Visual results achieved by Tophat\cite{4-tophat}, IPI\cite{10-IPI}, RIPT\cite{12-RIPT}, and our DNA-Net. The \textcolor{red} {correctly detected targets} and \textcolor{orange} {false alarms} are highlighted by red and orange dotted circles, respectively.}\label{Fig_1}
\end{figure}

To detect infrared small targets, numerous traditional methods have been proposed, including filtering-based methods \cite{4-tophat,5-maxmedian}, local-contrast-based methods \cite{6-lcm,7-Robust-lcm,8-TLLCM,9-WSLLCM,Local_contrast_01,Local_contrast_02}, and low-rank-based methods \cite{10-IPI,11-NRAM,12-RIPT,13-PSTNN,low_rank_01,low_rank_02}.
However, these traditional methods heavily rely on handcrafted features. Considering the characteristics of real scenes (e.g., target size, target shape, SCR, and clutter background) change dramatically, it is difficult to use handcrafted features and fixed hyper-parameters to handle such variations.

Different from traditional methods, CNN-based methods can learn features of infrared small targets in a data-driven manner. Liu et al. \cite{18-five-layer} proposed the first CNN-based SIRST detection method. They designed a multi-layer perception (MLP) network with 5 layers for infrared small target detection. Then, McIntosh et al. \cite{19-infrared-car} fine-tuned several existing generic object detection networks (e.g., Faster-RCNN \cite{20-faster-rcnn} and Yolo-v3 \cite{21-yolov3}) for infrared small target detection.  Specifically, Dai et al. \cite{22-ACM} proposed the first segmentation-based SIRST detection method. They designed an asymmetric contextual module (ACM) to replace the plain skip connection of Unet \cite{15-Unet}.
Although recent CNN-based methods have achieved the state-of-the-art performance, most of them only fine-tuned these networks designed for generic objects. Since the size of infrared small targets
is much smaller than generic objects, directly applying these methods for SIRST detection can easily lead to the loss of small targets in deep layers.

\begin{figure}
\centering
\includegraphics[width=8.8cm]{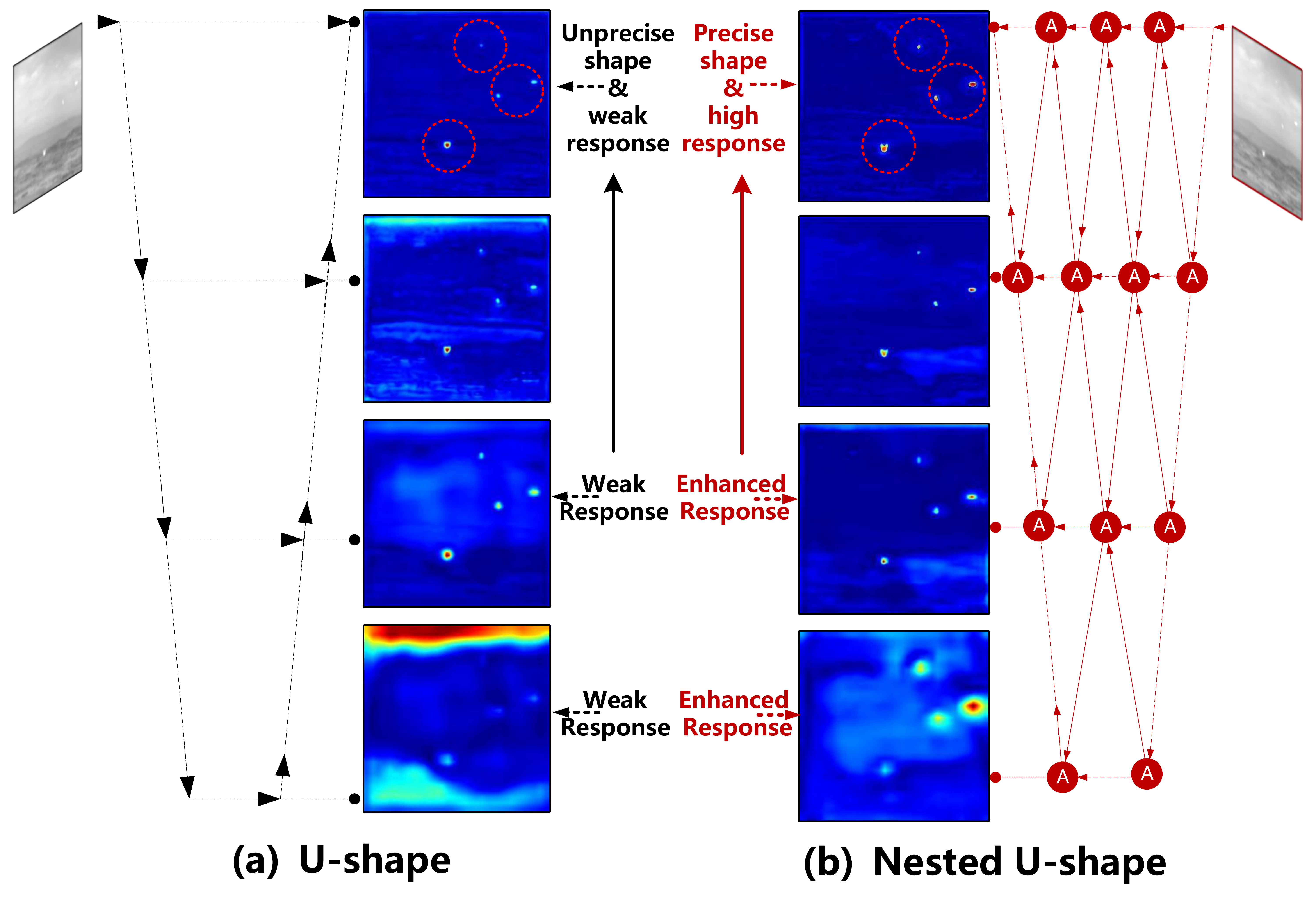}
\caption{The representation of small targets in deep CNN layers of (a) U-shape network (b) our Dense Nested U-shape (DNA-Net) network.}\label{Fig_22}
\end{figure}

Inspired by the success of nested structure in  medical image segmentation  \cite{Mdu-net,DMPU-net,Unet_3+, 25-Unet++} and hybrid attention in generic object detection \cite{CBAM}, we propose a dense nested attention network (namely, DNA-Net) to maintain small targets in deep layers. Specifically, we design a tri-directional dense nested interactive module (DNIM) with a cascaded channel and spatial attention module (CSAM) to achieve progressive feature interaction and adaptive feature enhancement. Within our DNIM, multiple nodes are imposed on the pathway between the encoder and decoder sub-networks. As shown in Fig.~\ref{Fig_22}(b), all nodes in our network are connected with each other to form a nested-shape network. Using DNIM, those middle nodes can receive features from their own and the adjacent two layers, leading to repetitive multi-layer feature fusion at deep layers.
Through repetitive feature fusion and enhancement, our network can maintain the targets in deep layers. Meanwhile, contextual information of maintained small targets can be well incorporated and fully exploited. Otherwise, as shown in Fig.~\ref{Fig_22}(a), the traditional U-shape network suffers from the loss of small targets in deep layers, which ultimately leads to inferior performance. In addition, we develop a novel infrared small target dataset (namely, the NUDT-SIRST dataset) to evaluate the performance of SIRST detection methods under different clutter backgrounds, target shapes, and target sizes. In summary, the contributions of this paper can be summarized as follows.

\begin{itemize}
\item 	We propose a DNA-Net to maintain small targets in deep layers. The contextual information of small targets can be well incorporated and fully exploited by repetitive feature fusion and enhancement.
\item 	A dense nested interactive module and a channel-spatial attention module are proposed to achieve progressive feature fusion and adaptive feature enhancement.
\item 	We develop an infrared small target dataset (namely, NUDT-SIRST). To the best of our knowledge, our dataset is the largest dataset with numerous categories of target shapes, various target sizes, diverse clutter backgrounds, and ground truth annotations.
\item Experiments on both public and our NUDT datasets demonstrate the superior performance of our method. Compared to existing methods, our method is more robust to the variations of clutter background, target size, and target shape (as shown in Fig.~\ref{Fig_1}).
\end{itemize}

\begin{figure*}
 \centering
 \includegraphics[width=18.2cm]{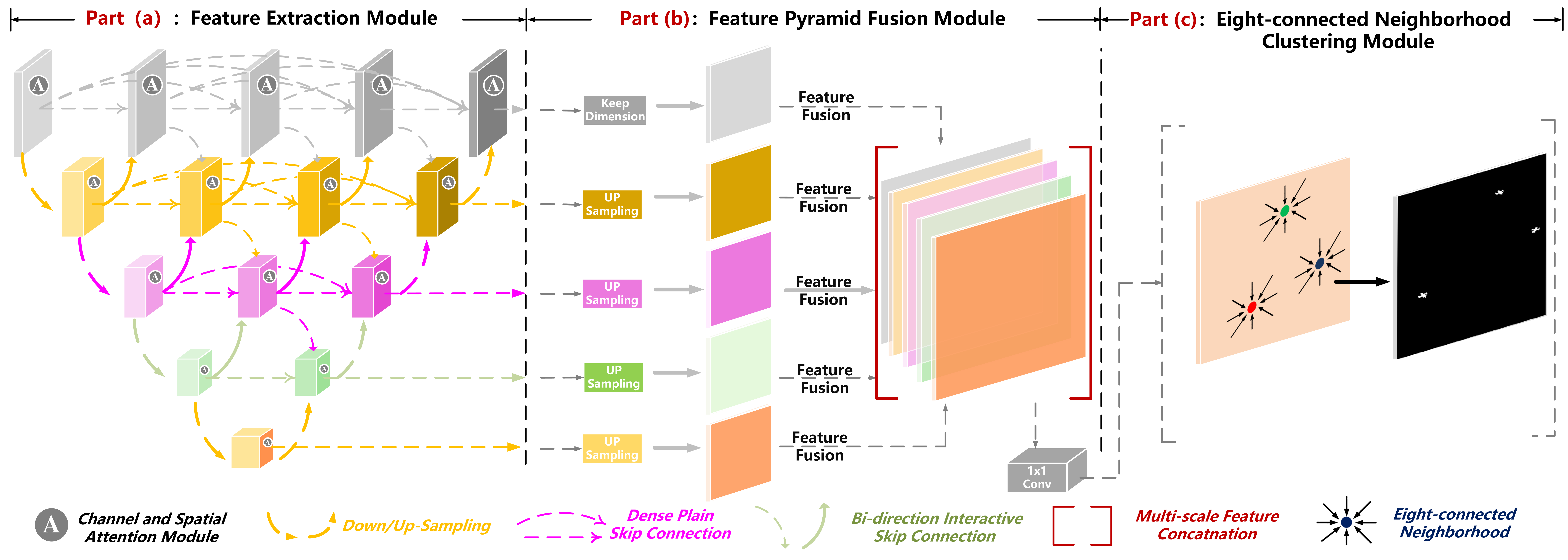}
\caption{An illustration of the proposed dense nested attention network (DNA-Net). (a) Feature extraction module. Input images are first fed into the dense nested interactive module (DNIM) to aggregate information from multiple scales. Note that, features from different semantic levels are adaptively enhanced by a channel and spatial attention module (CSAM). (b) Feature pyramid fusion module (FPFM). The enhanced features are upsampled and concatenated to achieve multi-layer output fusion. (c) Eight-connected neighborhood clustering algorithm. The segmentation map is clustered to determine the centroid of each target region. }\label{4_network}
 \end{figure*}

This paper is organized as follows: In Section \ref{SecRelatedWork}, we briefly review the related work. In Section \ref{SecMethodology}, we introduce the architecture of our DNA-Net and our self-developed dataset in details. In Section \ref{dataset}, we introduce our self-developed NUDT-SIRST dataset in details. The experimental results are represents in Section \ref{SecExperiment}. Section \ref{SecConclusion} gives the conclusion.

\section{Related Work}\label{SecRelatedWork}

In this section, we briefly review the major works in SIRST detection and corresponding datasets.

\subsection{Single-frame Infrared Small Target Detection}

SIRST detection has been extensively investigated for decades. The traditional paradigm achieves SIRST detection by measuring the discontinuity between targets and backgrounds. Typical methods include filtering-based methods \cite{4-tophat,5-maxmedian}, local contrast measure based methods \cite{6-lcm,7-Robust-lcm,8-TLLCM,9-WSLLCM,Local_contrast_01,Local_contrast_02}, and low rank based methods \cite{10-IPI,11-NRAM,12-RIPT,13-PSTNN,low_rank_01,low_rank_02}. Considering real scenes are much more complex with dramatic changes target size, shape, and clutter background, it is difficult to use handcrafted features and fixed hyper-parameters to handle such variations.
To address this problem, recent CNN-based methods learn trainable features in a data-driven manner. Thanks to the large quantity of data and the powerful model fitting capability of CNNs, these methods achieve better performance than traditional ones.

Existing CNN-based methods can be divided into detection based methods and segmentation based methods. Liu et al. \cite{18-five-layer} first introduced a generic target detection framework for infrared small target detection. They designed a multi-layer perception (MLP) network with 5 layers for infrared small target detection. Then, McIntosh et al. \cite{19-infrared-car} fine-tuned several generic target detection network (e.g., Faster-RCNN \cite{20-faster-rcnn} and Yolo-v3 \cite{21-yolov3}) and used the optimized eigen-vectors as input to achieve improved performance.

Recently, segmentation-based methods have attracted increasing attention. That is because, these methods can produce both pixel-level classification and localization outputs.
Dai et al. \cite{22-ACM} proposed the first segmentation-based network (i.e., ACM). They designed an asymmetric contextual module to aggregate features from shallow layers and deep layers. Then, Dai et al. \cite{23-ALCNet} further improved their ACM by introducing a dilated local contrast measure. Specifically, a feature cyclic shift scheme was designed to achieve a trainable local contrast measure. Moreover, Wang et al. \cite{24-ICCV19} decomposed the infrared target detection problem into two opposed sub-problems (i.e., miss detection and false alarm) and used a conditional generative adversarial network (CGAN) to achieve the trade-off between miss detection and false alarm for infrared small target detection.

Although the performance is continuously improved by recent networks, the loss of small targets in deep layers still remains. This problem ultimately results in the poor robustness to dramatic scene changes (e.g., clutter background, targets with different SCR, shape, and size).

\subsection{Datasets for SIRST Detection}

 Existing open-source dataset in infrared small target detection is scarce, most traditional methods are evaluated on their in-house datasets. Only a few infrared small target datasets are released by CNN-based methods \cite{24-ICCV19,22-ACM}. Wang et al.\cite{24-ICCV19} built the first big and open SIRST dataset. This dataset includes 10000 training images and 100 test images. However, many targets in this dataset do not meet the definition of society of photo-optical instrumentation engineers (SPIE) \cite{SPIE} and have obvious synthesized traces with illogical annotations. These problems may lead to the inapplicability toward SIRST detection. Dai et al.\cite{22-ACM} built the first real SIRST dataset with high-quality images and labels. However, the number of images in NUAA-SIRST is 427 (256 for training), which cannot well cover dramatic scene changes in infrared small target detection. Moreover, these real infrared data are all manually labelled with many inaccurately labeled pixels.

Although these open-sourced datasets greatly prompt the prosperity of SIRST detection, their limited data capacity, data variety, and poor annotation hinder the further development of this field. Synthesized data can be easily generated to achieve higher variety and annotation quality at very low cost (i.e., time and money). Hence, we developed a new NUDT-SIRST dataset with numerous categories of target, vairous target sizes, diverse clutter backgrounds, and accurate annotations. The superiority of our dataset is evaluated in Section \ref{SecExperiment}.

\section{Methodology}\label{SecMethodology}

In this section, we introduce our DNA-Net in details.

\subsection{Overall Architecture}\label{SecOverall}

As illustrated in Fig.~\ref{4_network}, our DNA-Net takes a SIRST image as its input and sequentially performs feature extraction (Section \ref{SecDNM}), feature pyramid fusion (Section \ref{SecFPM}), and eight-connected neighborhood clustering (Section \ref{SecCluster}) to generate the detection results.

Section \ref{SecDNM} introduces the motivation of our feature extraction module and the architecture of the dense nested interactive module (DNIM) and the channel-spatial attention module (CSAM). Input images are first preprocessed and fed into the backbone of DNIM to extract multi-layer features. Then, multi-layer features are repetitively fused at the middle convolution nodes of skip connection and then are gradually passed into the decoder subnetworks. Due to the semantic gap at multi-layer feature fusion stage of DNIM, we used CSAM to adaptively enhance these multi-level features for achieving better feature fusion. Section \ref{SecFPM} presents the feature pyramid fusion module. Enhanced multi-layer features at each scale are upscaled to the same size. Next, the shallow-layer features with rich spatial information and deep-layer features with high-level information are concatenated to generate robust feature maps. Section \ref{SecCluster} elaborates the eight-connected neighborhood clustering module. Feature maps are fed into this module to calculate the spatial location of target centroid, which is then used for comparison in Section \ref{SecExperiment}.

\subsection{The Feature Extraction Module}\label{SecDNM}

\subsubsection{Motivation}

 \begin{figure}
\centering
\includegraphics[width=8.8cm]{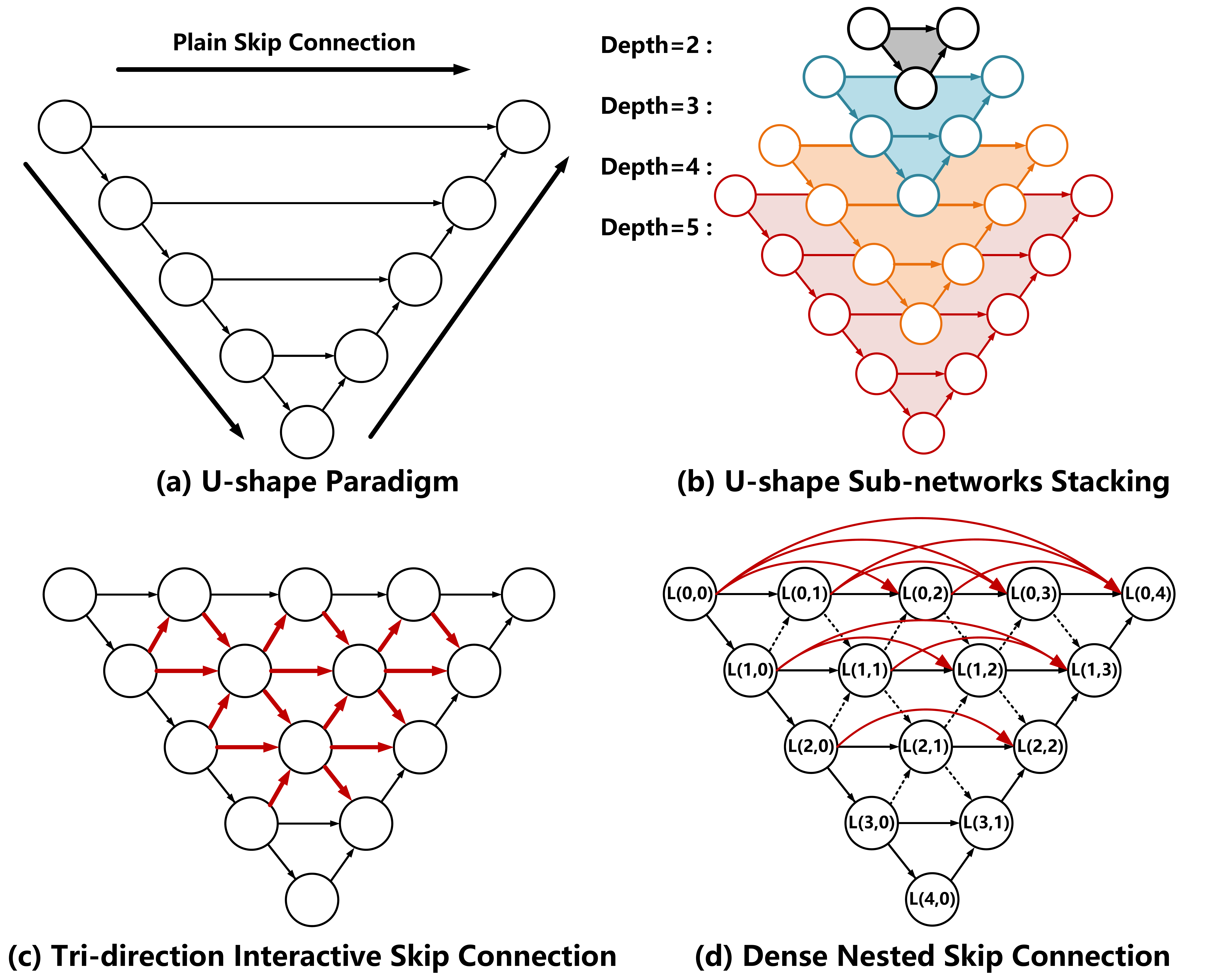}
\caption{An illustration of the U-shape structure and our dense nested structure. The insight comes from the multiple U-shape subnetwork stacking. The representation of small targets in the deep layers is maintained and the high-level information is extracted.}\label{5_Network_inference}
\end{figure}

As shown in Fig.~\ref{5_Network_inference}(a), traditional U-shape structure\cite{15-Unet} consists of an encoder, a decoder, and plain skip connections. The encoder is used to enlarge the receptive field and extract high-level information. Decoder helps to recover the size of feature maps (which finally reach the same size as the input images) and achieve progressive multi-scale feature fusion. The plain skip connection acts as a bridge to pass these low-level and high-level features from encoder to decoder subnetworks.

To achieve powerful contextual information modeling capability, a straightforward way is to continuously increase the number of layers. In this way, high-level information can be obtained and larger receptive field can be achieved. However, infrared small targets are significantly different in their sizes, ranging from one pixel (i.e., point targets) to tens of pixels (i.e., extended targets). With the increase of network layers, high-level information of extended targets is obtained, while the point targets are easily lost after multiple pooling operation. Therefore, we should design a special module to extract high-level features and maintain the representation of small targets in the deep layers.

\subsubsection{The Dense Nested Interactive Module}

 As shown in Fig.~\ref{5_Network_inference}(b), we stack multiple U-shape sub-networks together to build a dense nested structure. Since the optimal receptive field for different sizes of targets varies a lot, these U-shape sub-networks with different depths are naturally suitable for targets with different sizes. Based on this idea, we impose multiple nodes in the pathway between encoder and decoder sub-networks. All of these middle nodes are densely connected with each other to form a nested-shape network. As shown in Fig.~\ref{5_Network_inference}(c) and (d), each node can receive features from its own and the adjacent layers, leading to repetitive multi-layer feature fusion. As a result, the representations of small targets are maintained in the deep layers and thus better results can be achieved.

In this paper, we stack \textit{I} layers of DNIM to form our feature extraction module. Without loss of generality, we take the $i^{th} (i=0,1,2,...,I)$ DNIM layer as an example to introduce this structure, as shown in Fig.~\ref{5_Network_inference}(c) and (d). Assume $\mathbf{L}^{i,j}$ denote the output of node $\hat{\mathbf{L}}^{i,j}$, where $i$ is the $i^{th}$ down-sampling layer along the encoder and $j$ is the $j^{th}$ convolutional layer of dense block along the plain skip pathway. When $j=0$ , each node only receives features from dense plain skip connection. The stack of feature maps represented by $\mathbf{L}^{i,j}$ are computed as:
\begin{equation}\label{DNIM_1}
{\mathbf{L}}^{i,j}={\cal{P}}_{max}({\cal{F}}(\mathbf{L}^{i-1,j})),
\end{equation} where $\cal{F}(\cdot)$ denotes multiple cascaded convolution layers of the same convolution block. ${\cal{P}}_{max}(\cdot)$ denotes max-pooling with a stride of 2.
When $j>0$ , each node receives outputs from three directions including dense plain skip connection and nested bi-direction interactive skip connection, the stack of feature maps represented by $\mathbf{L}^{i,j}$ is generated as:

\begin{equation}\label{DNIM_2}
{\mathbf{L}}^{\textit{i},\textit{j}}=\Big[{\cal{F}}\left[\mathbf{L}^{\textit{i},\textit{k}}\right]^{\textit{j}-1}_{\textit{k}=0},{\cal{P}}_{\textit{max}}({\cal{F}}(\mathbf{L}^{\textit{i}+1,\textit{j}-1})), {\cal{U}}({\cal{F}}(\mathbf{L}^{\textit{i}-1,\textit{j}})) \Big],
\end{equation} where $\cal{U}(\cdot)$ denotes the up-sampling layer, and $\left[\;\cdot,\cdot\right]$ denotes the concatenation layer.

\subsubsection{Channel and Spatial Attention Module}
\begin{figure}
\centering
\includegraphics[width=8.8cm]{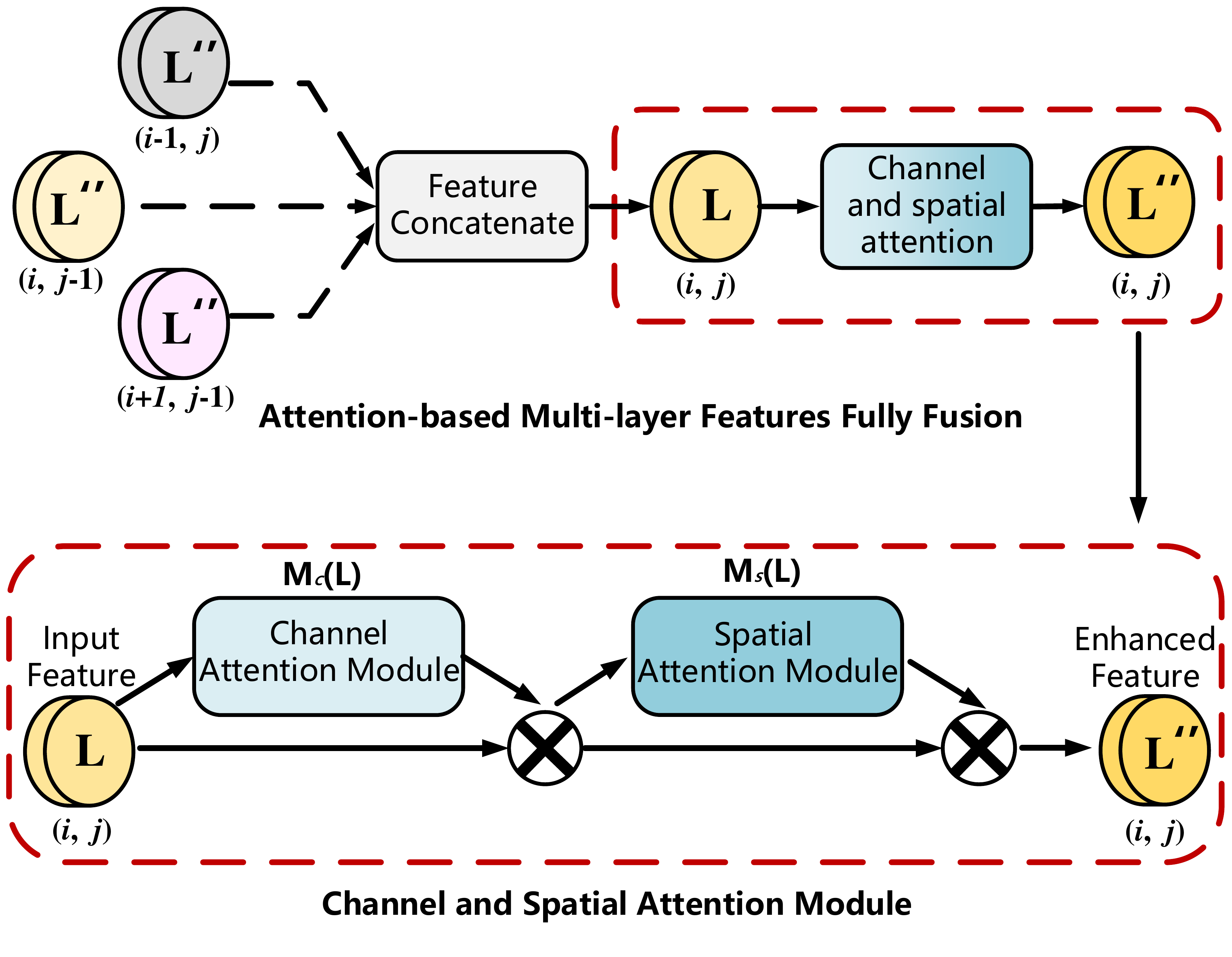}
\caption{ Channel and spatial attention module. CSAM is used to reduce the semantic gap at the multi-layer feature fusion stage in DNIM.}\label{6_SCAM}
\end{figure}

 As shown in Fig.~\ref{6_SCAM}, CSAM is used for adaptive feature enhancement after each multi-layer feature fusion of DNIM.

\begin{table*}[]
\centering
\renewcommand\arraystretch{1.1}

\caption{Main characteristics of several popular SIRST datasets. Note that, our NUDT-SIRST dataset contains common background scenes, various target types, and the most ground truth annotations.} \label{TabOVER_dataset}
\label{DiagnosisResults}
\begin{tabular}{|l|c|c|c|c|c|c|}
\hline
Datasets   & Image Type      & Background   Scene         & \#Image & Label   Type            & Target   Type    & Public \\ \hline
NUAA-SIRST(ACM)\cite{22-ACM} & real      & Cloud$/$City$/$Sea                 & 427     & Manual   Coarse Label       & Point/Spot/Extended &$\surd$   \\ \hline
NUST-SIRST\cite{24-ICCV19} & synthetic & Cloud$/$City$/$River$/$Road        & 10000   & Manual   Coarse Label       & Point/Spot  &$\surd$ \\ \hline
CQU-SIRST(IPI)\cite{10-IPI}  & synthetic & Cloud$/$City$/$Sea                 & 1676     & Ground Truth                & Point/Spot      &$\times$    \\ \hline
NUDT-SIRST(ours) & synthetic & Cloud$/$City$/$Sea$/$Field$/$Highlight & 1327    & Ground Truth                & Point/Spot/Extended &$\surd$\\ \hline
\end{tabular}
\end{table*}

\begin{figure*}[htbp]
    \centering
    \subfigure[the number of targets]{
    \begin{minipage}[t]{0.33\linewidth}
    \centering
    \includegraphics[width=6.0 cm]{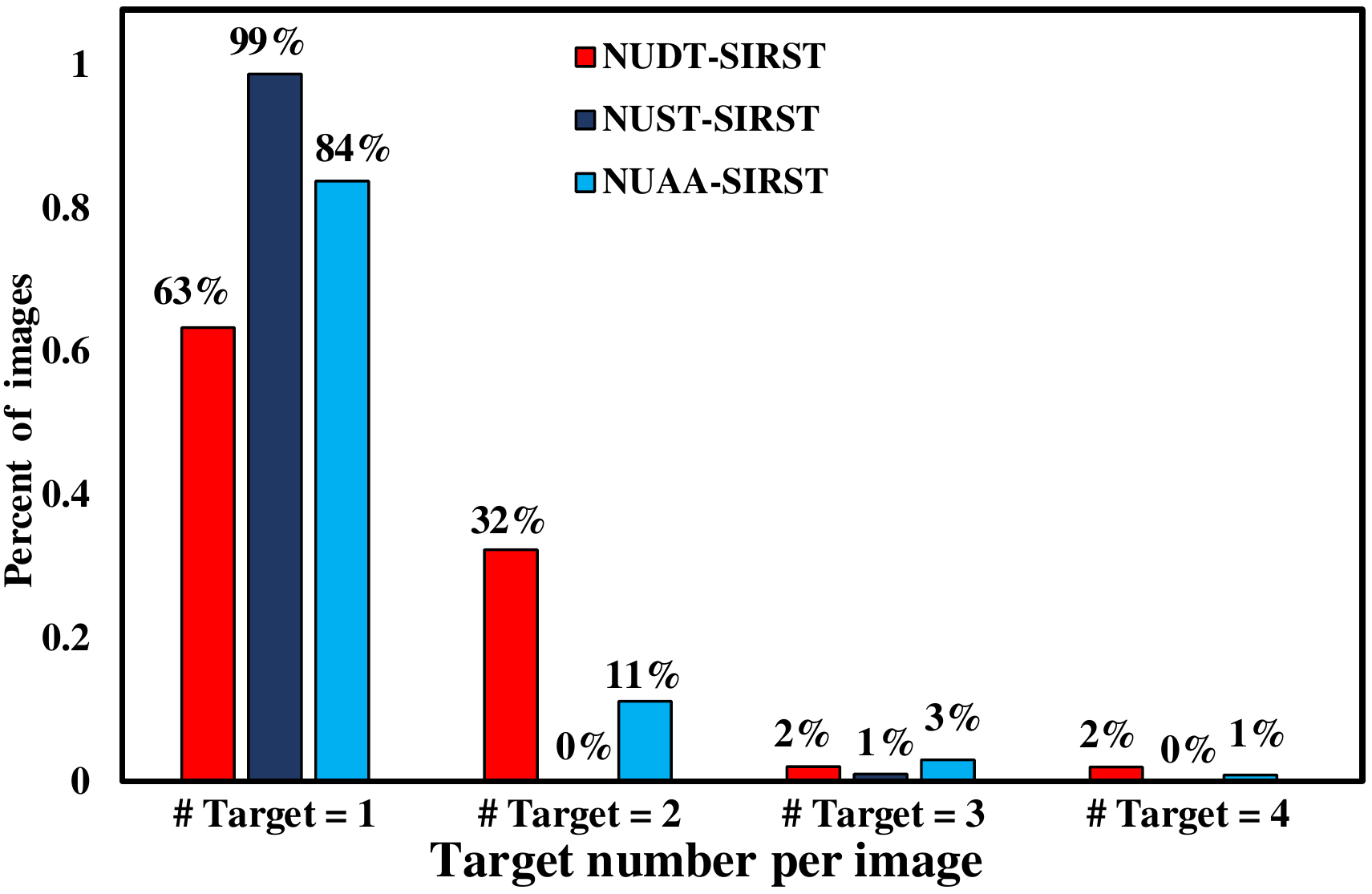}
    \end{minipage}%
    }%
    \subfigure[target size]{
    \begin{minipage}[t]{0.33\linewidth}
    \centering
    \includegraphics[width=6.0 cm]{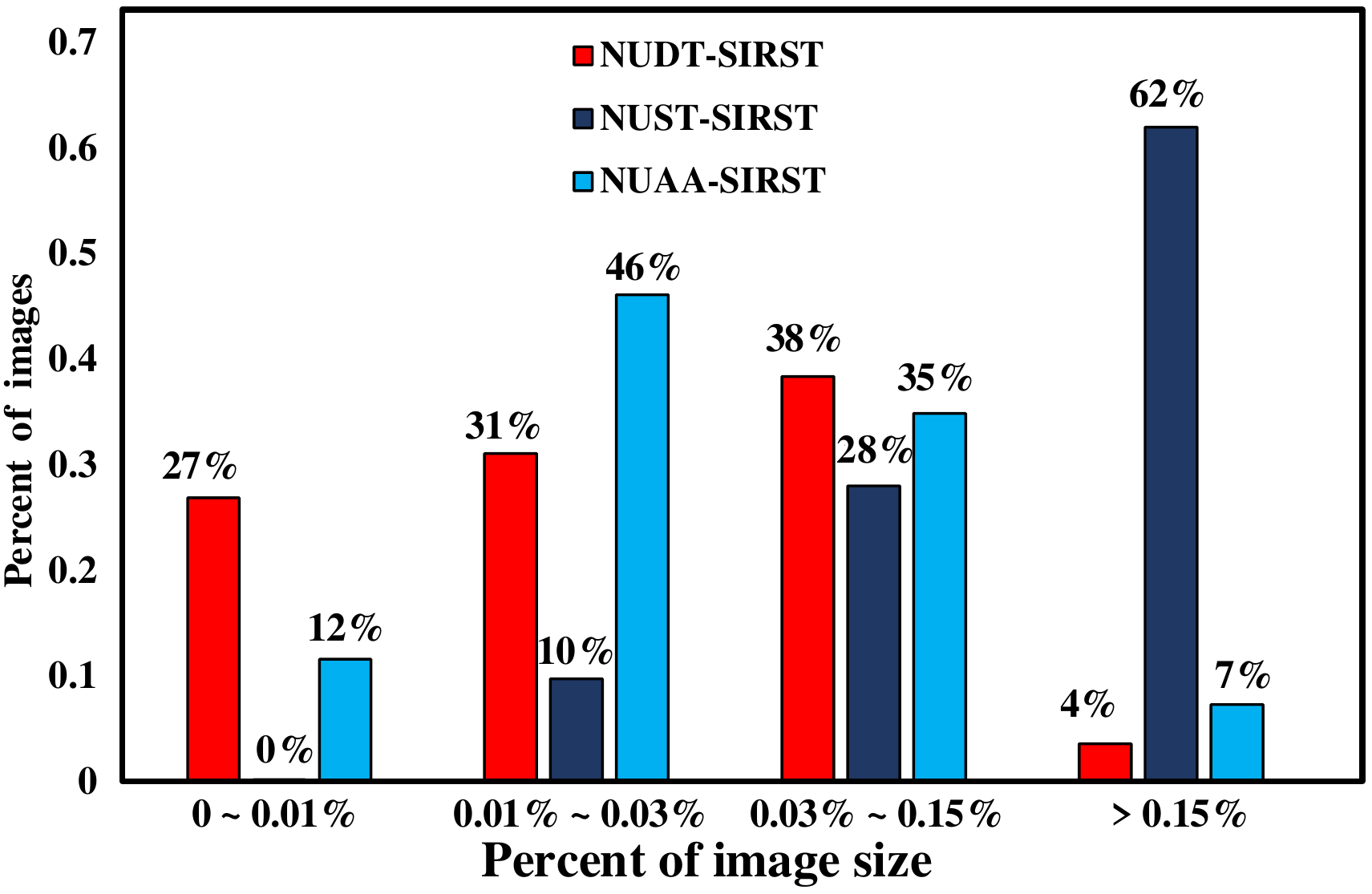}
    \end{minipage}%
    }%
    \subfigure[target brightness]{
    \begin{minipage}[t]{0.33\linewidth}
    \centering
    \includegraphics[width=6.0 cm]{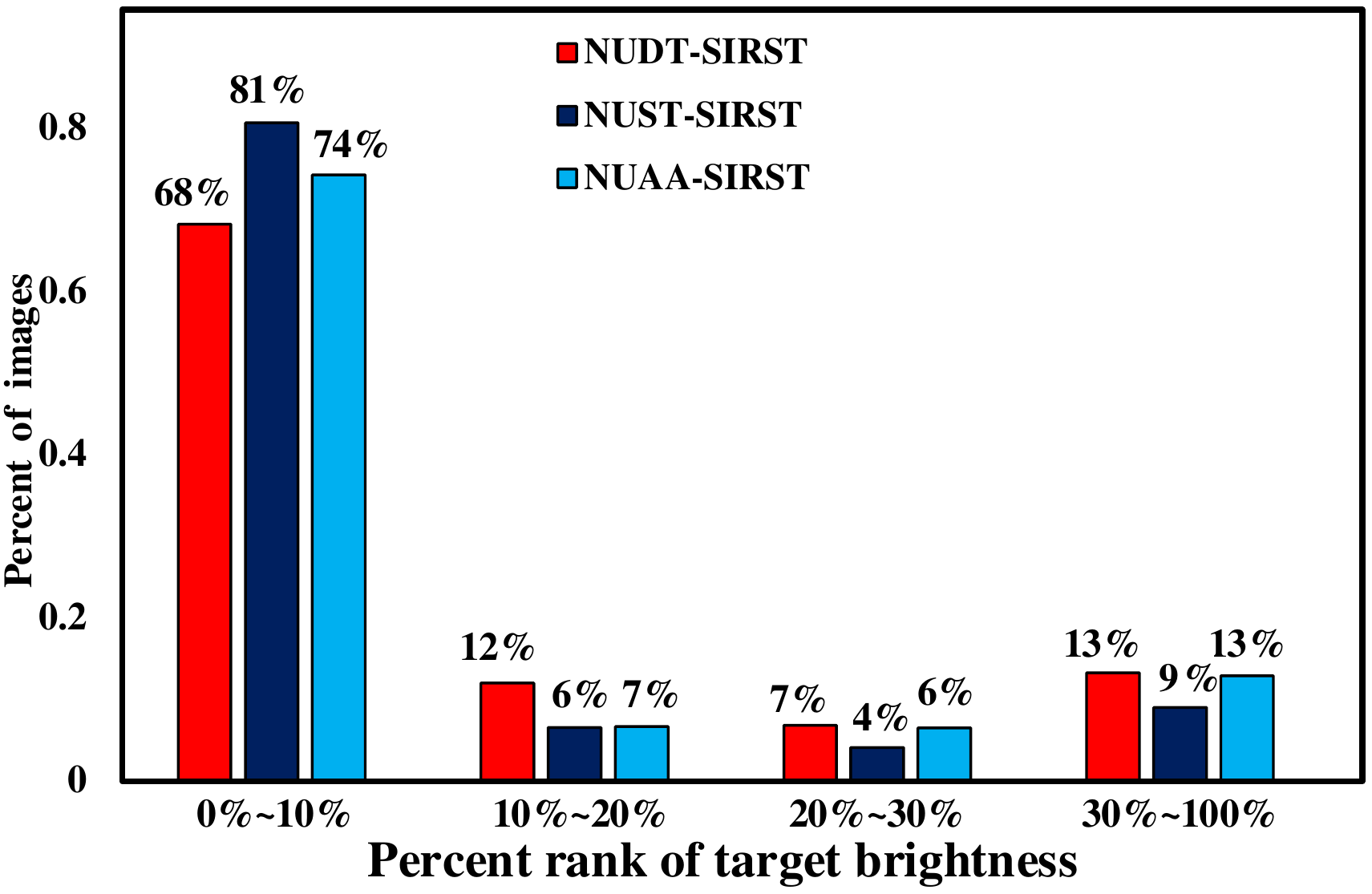}
    \end{minipage}%
    }%
    \caption{Comparison of existing public SIRST datasets on (a) the number of targets, (b) target size, and (c) target brightness. Our NUDT-SIRST dataset contains more multi-target scenarios, more small targets, and less visually salient targets.}\label{Dataset_evaluation}
    \end{figure*}

The CSAM consists of two cascaded attention units. The feature maps $\mathbf{L}^{i,j}$ from node $\hat{\mathbf{L}}^{i,j}$ ($i\in\{0,1,2,...I\}$, $j\in\{0,1,2,...J\}$) are sequentially processed by a 1D channel attention map $\mathbf{M}_{c}\in\mathbb{R}^{{C}_{i}\times{1}\times{1}}$ and a 2D spatial attention map $\mathbf{M}_{s}\in\mathbb{R}^{{1}\times{H}_{i}\times{W}_{i}}$. The channel attention process can be summarized as follows:
\begin{equation}\label{SCAM_1}
\mathbf{M}_{c}(\mathbf{L})=\sigma\Big[MLP({\cal{P}}_{max}(\mathbf{L})) + (MLP({\cal{P}}_{avg}(\mathbf{L}))\Big],
\end{equation}

\begin{equation}\label{SCAM_1}
\mathbf{L}^{'}=\mathbf{M}_{c}(\mathbf{L})\otimes\mathbf{L},
\end{equation}where $\otimes$ denotes the element-wise multiplication, $\sigma$ denotes sigmoid function, ${C}_{i},{H}_{i},{W}_{i}$ denote the number of channels, height, and width of $\mathbf{L}^{i,j}$. ${\cal{P}}_{avg}(\cdot)$ denotes average pooling with a stride of 2, respectively. The shared network is composed of a multi-layer perceptron (MLP) with one hidden layer. Before multiplication, the attention maps $\mathbf{M}_{c}(\mathbf{L})$ are stretched to the size of $\mathbf{M}_{c}(\mathbf{L})\in\mathbb{R}^{{C}_{i}\times{H}_{i}\times{W}_{i}}$.

Similar to channel attention process, the spatial attention process can be summarized as follows:

\begin{equation}\label{SCAM_4}
\mathbf{M}_{s}(\mathbf{L}^{'})=\sigma\Big[f^{7\times7}({\cal{P}}_{max}(\mathbf{L}^{'})), ({\cal{P}}_{avg}(\mathbf{L}^{'})\Big],
\end{equation}

\begin{equation}\label{SCAM_3}
\mathbf{L}^{''}=\mathbf{M}_{s}(\mathbf{L}^{'})\otimes\mathbf{L}^{'},
\end{equation}where $f^{7\times7}$ represents a convolutional operation with the filter size of 7$\times$7. The attention maps $\mathbf{M}_{s}(\mathbf{L})$ are also stretched to the size of $\mathbf{M}_{c}(\mathbf{L})\in\mathbb{R}^{{C}_{i}\times{H}_{i}\times{W}_{i}}$ before multiplication.

\subsection{The Feature Pyramid Fusion Module}\label{SecFPM}
After the feature extraction module, we develop a feature pyramid fusion module to aggregate the resultant multi-layer features. As shown in Fig.~\ref{4_network} (b), we first upscale multi-layer features to the same size of $\mathbf{L}^{i,J}_{en\_up}\in\mathbb{R}^{{C}_{i}\times{H}_{0}\times{W}_{0}}$ $i\in\{0,1,...,I\}$. Then, the shallow-layer feature with rich spatial and profile information and deep-layer feature with rich semantic information are concatenated to generate global robust feature maps:

\begin{equation}\label{FPM_1}
\mathbf{G}=\{\mathbf{L}^{0,J}_{en\_up},\mathbf{L}^{1,J}_{en\_up},...,\mathbf{L}^{I,J}_{en\_up}\}.
\end{equation}

\subsection{The Eight-connected Neighborhood Clustering Module}\label{SecCluster}

After the feature pyramid fusion module, we introduce an eight-connected neighborhood clustering module \cite{cluster} to cluster the pixels belonging to the same target together and calculate the centroid of each target. If any two pixels ${(m_{0},n_{0})}$, ${(m_{1},n_{1})}$ in feature maps $\mathbf{G}$ have intersection areas in their eight neighborhoods, i.e.,

\begin{equation}\label{ENCM_2}
{{\cal{N}}_{8}}_{(m_{0},n_{0})}\cap{{\cal{N}}_{8}}_{(m_{1},n_{1})}\ \ne \varnothing,
\end{equation}where ${{\cal{N}}_{8}}_{(m_{0},n_{0})}$ and ${{\cal{N}}_{8}}_{(m_{1},n_{1})}$ represent the eight neighborhoods of pixel ${(m_{0},n_{0})}$ and ${(m_{1},n_{1})}$, ${(m_{0},n_{0})}$ and ${(m_{1},n_{1})}$ are judged as adjacent pixels. Then, if the these two pixels have the same value (0 or 1), i.e.,

\begin{equation}\label{ENCM_3}
\mathbf{g}_{(m_{0},n_{0})}=\mathbf{g}_{(m_{1},n_{1})}, \forall \mathbf{g}_{(m_{0},n_{0})},\mathbf{g}_{(m_{1},n_{1})}\in \mathbf{G},
\end{equation} where $\mathbf{g}_{(m_{0},n_{0})}$ and $\mathbf{g}_{(m_{1},n_{1})}$ represent the gray value of pixel  ${(m_{0},n_{0})}$ and ${(m_{1},n_{1})}$, these two pixels are considered to be in a connected area. Pixels in a connected area belong to the same targets. Once all targets in the image are determined, centroid can be calculated according to their coordinate.

\begin{figure*}
\centering
\includegraphics[width=16.2cm]{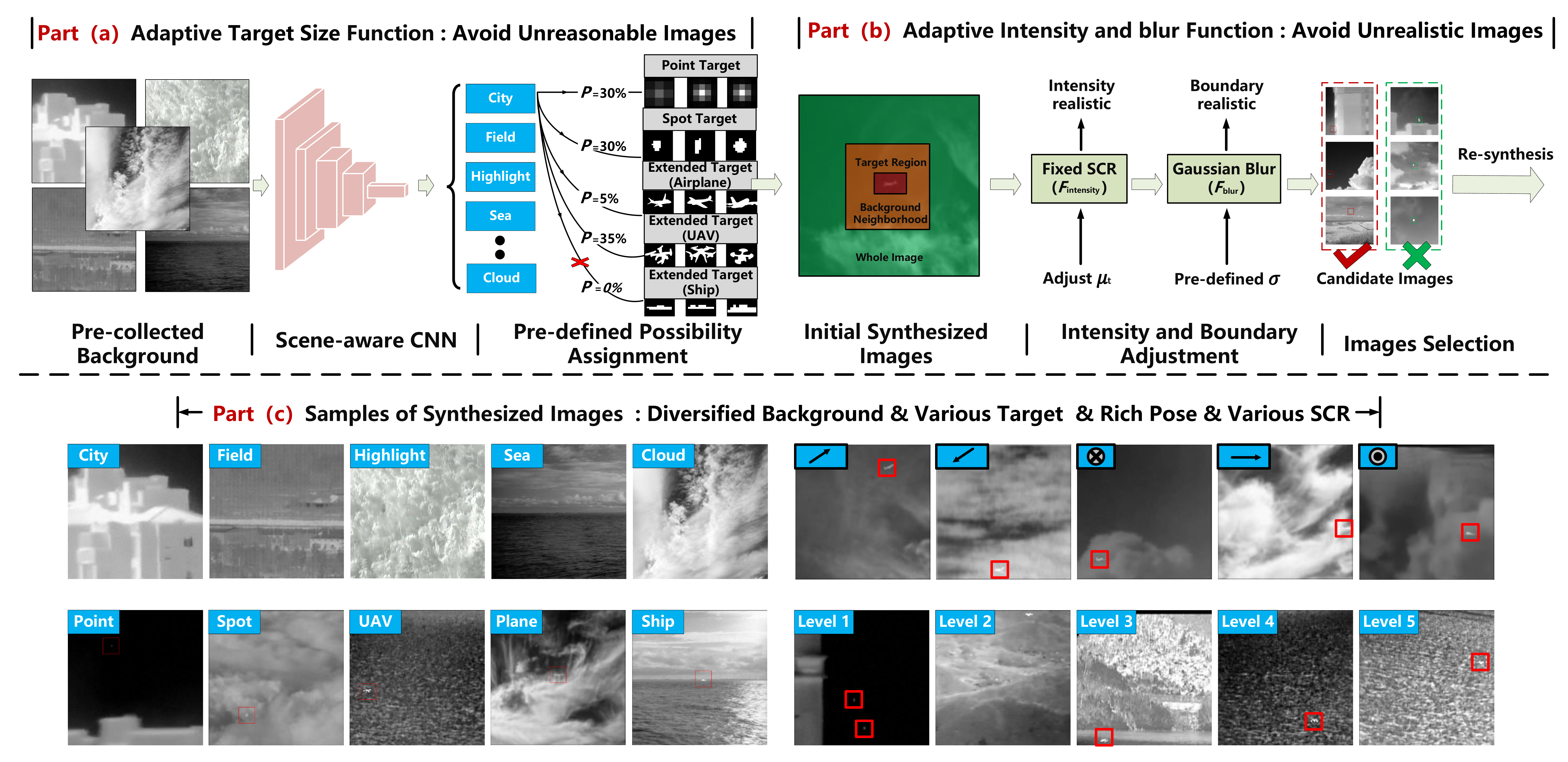}
\caption{Synthesis process of our dataset. (a) Adaptive target size function. Pre-collected background images are fed into a scene-aware CNN model to identify the type of background. Then, the size and type of candidate targets are selected with pre-defined possibility $\emph{P}_{size}$. The background and selected targets are directly added. (b) Adaptive intensity and blur function. The initially synthesized images are sequentially fed into an adaptive intensity function $\mathbf{F}_{intensity}$ and a Gaussian blur function $\mathbf{F}_{blur}$ to make the targets' intensity and boundary realistic, respectively. (c) Samples of our NUDT-SIRST dataset. Our dataset covers multiple real infrared backgrounds, various target types, rich poses, and ground truth labels. $\nearrow$, $\swarrow$, $\rightarrow$, and $\otimes$ represents different moving directions of targets.}\label{Fig_generate_process}
\end{figure*}

\section{The NUDT-SIRST dataset}\label{dataset}
\subsection{Motivation}\label{SecMotivation}

Quality, quantity, and scene diversity of data significantly affect the performance of CNN-based methods.  As shown in Table~\ref{TabOVER_dataset}, existing datasets either lack enough scenes (e.g., NUST-SIRST \cite{24-ICCV19} and CQU-SIRST \cite{10-IPI}) or have limited data capacity (e.g., NUAA-SIRST \cite{22-ACM}). It is costly to collect a large-scale dataset with accurate pixel-level annotations. These issues hinder the further development of CNN-based methods. Inspired by the solutions in other data-scarcity field (e.g., ship detection \cite{17-Shipdetection,zhangfeng}, moving car detection \cite{xiaochao,Qianyin}), we develop a large-scale infrared small target dataset (namely, the NUDT-SIRST dataset). Our NUDT-SIRST dataset enables performance evaluation of CNN-based methods under numerous categories of target type, target size, and diverse clutter backgrounds. As shown in Fig.~\ref{Fig_generate_process}(c), our dataset contains 5 main background scenes including city, field, highlight, sea, and cloud. Each image is synthesized from real background with various targets (e.g., point, spot, and extended) under various SCR and rich poses. Note that, most of the background images are collected by ourselves, only a few field-type background images are adopted from \cite{ART}. The detailed synthesis process and comparison among datasets are introduced in Section \ref{SecDetails} and Section \ref{SecComparison}.

\subsection{Implementation Details}\label{SecDetails}

High-quality synthesized images should be both physically reasonable and visually realistic. To render reasonable images, as shown in Fig.~\ref{Fig_generate_process}(a), we first used a Gaussian kernel function and collected target templates (e.g., spot, plane, ship, and UAV) to simulate point, spot, and extended targets, respectively. Then, we adopted an adaptive target size function  $\mathbf{F}_{size}$ to make sure the size of target and the combination of virtual targets with real infrared background reasonable. In this function, a scene-aware CNN $\mathbf{F}_{scene}$ is first used to identify the type of the background. Then, we assigned pre-defined possibility  $\emph{P}_{size}$ to identify the size and type of candidate targets. In this way, we can avoid the unreasonable combination of target and background such as a big plane target with city background and a ship target with sky background.

To generate visually realistic images, as shown in Fig.~\ref{Fig_generate_process}(b), we used an adaptive intensity function $\mathbf{F}_{intensity}$ and a Gaussian blur function $\mathbf{F}_{blur}$ to adjust the target's intensity and blur it's boundary, respectively. In the adaptive intensity function, we adjusted the average gray value of the target to keep the target's SCR fixed at an empirical value $\emph{C}$ (i.e., 3, 4, 5, and 6). That is:

 \begin{equation}\label{Magnitude_function}
\emph{SCR}=\left| \frac{{\mu}_{T}-{\mu}_{B}}{{\sigma}_{B}} \right|=\emph{C},
\end{equation} where ${\mu}_{B}$ and ${\sigma}_{B}$ are the average and standard derivation of the background. Then, we imposed a $5\times5$ Gaussian blur function with different $\sigma$ (i.e., 0.2, 0.5, 1.0, etc.) on the images to ensure the smoothness of the synthesized images. Finally, we manually removed visually low-quality images.

 \subsection{Comparison to Existing Datasets}\label{SecComparison}
In this subsection, we compare our NUDT-SIRST dataset to several public SIRST datasets. Following \cite{22-ACM}, we use three metrics (i.e., the number of targets, target size, and target brightness) to evaluate these datasets. As shown in Fig.~\ref{Dataset_evaluation}(a), about 37\% of images in the NUDT-SIRST dataset contain no less than 2 targets. This ratio is much higher than the other two datasets. Target size distribution in Fig.~\ref{Dataset_evaluation}(b) shows that 27\% of targets occupy no more than 0.01\% area of the whole image and 96\% of targets meet the SPIE's defination for small targets (i.e., the target should be smaller than 0.15\% area of the whole image). Point and small target ratios are much higher than the other two datasets. As shown in Fig.~\ref{Dataset_evaluation}(c), there are about 32\% of targets locating outside of top 10\% of the image brightness value. It demonstrates that the images of our dataset are less visually salient than other datasets. In summary, compared with existing datasets  \cite{22-ACM} \cite{24-ICCV19}, our dataset introduces more challenging scenes (i.e., multiple targets, point target, and dim target scenes).

\section{Experiment}\label{SecExperiment}
In this section, we first introduce our evaluation metrics and implementation details. Then, we compare our DNA-Net to several state-of-the-art SIRST detection methods. Finally, we present ablation studies to investigate our network.

\subsection{Evaluation Metrics}\label{Evaluation Metrics}

Pioneering CNN-based works\cite{22-ACM,23-ALCNet,24-ICCV19} mainly use pixel-level evaluation metrics like $IoU$, precision, and recall values. These  metrics mainly focus on the target shape evaluation. However, infrared small targets are generally lack of shapes and textures. For a $3\times3$ small target, one falsely predicted pixel will cause 11.1\% decrease in ${P}_{d}$. Consequently, these pixel-level evaluation metrics are unsuitable for small targets. Actually, the overall target localization is the most important criteria for SIRST detection. Therefore, we adopt ${P}_{d}$ and ${F}_{a}$ to evaluate the localization ability and use $IoU$ to evaluate shape description ability.

\begin{table*}[]
\centering
\renewcommand\arraystretch{1.2}
\caption{$IoU$, $P_{d}$, and $F_{a}$ values achieved by different state-of-the-art methods on the NUDT-SIRST and NUAA-SIRST datasets, For $IoU$ and $P_{d}$, larger values indicate higher performance. For $F_{a}$, smaller values indicate higher performance. The best results are in \textcolor{red} {red} and the second best results are in \textcolor{blue} {blue}.Tr=50\% means 50\% images are used for training and the rest are used for test.} \label{TabOVER_ALL}
\begin{tabular}{|l|c|c|c|c|c|c|c}
\hline
\multicolumn{1}{|c|}{\multirow{2}{*}{Method   Description}} & \multicolumn{3}{c|}{NUDT-SIRST (Tr=50\%)}         & \multicolumn{3}{c|}{NUAA-SIRST (Tr=50\%)}       \\ \cline{2-7}
\multicolumn{1}{|c|}{}   & \multicolumn{1}{c|}{$IoU$($\times10^{2}$)} & \multicolumn{1}{c|}{$P_{d}$($\times10^{2}$)} & \multicolumn{1}{c|}{$F_{a}$($\times10^{6}$)} & \multicolumn{1}{c|}{$IoU$($\times10^{2}$)} & \multicolumn{1}{c|}{$P_{d}$($\times10^{2}$)} & \multicolumn{1}{c|}{$F_{a}$($\times10^{6}$)}                     \\ \hline
Filtering   Based: Top-Hat\cite{4-tophat}                             &   20.72                &   78.41               & 166.7                 & 7.143                     & 79.84               & 1012                                        \\ \hline
Filtering   Based: Max-Median \cite{5-maxmedian}                      &   4.197             &     58.41             & 36.89                     & 4.172                     & 69.20                 & 55.33                                        \\ \hline
Local   Contrast Based: WSLCM \cite{9-WSLLCM}                         &   2.283                &       56.82             &   1309                & 1.158                     & 77.95                 & 5446                                       \\ \hline
Local   Contrast Based: TLLCM\cite{8-TLLCM}                           &   2.176                &     62.01              &  1608                 & 1.029                      & 79.09                 & 5899                                           \\ \hline
Low Rank   Based: IPI\cite{10-IPI}                                  &   17.76               &      74.49            &  41.23                  & 25.67                     & 85.55                 & 11.47                                            \\ \hline
Low Rank   Based: NRAM\cite{11-NRAM}                                &  6.927               &       56.40            &       19.27             & 12.16                   & 74.52                  & 13.85                                            \\ \hline
Low Rank   Based: RIPT\cite{12-RIPT}                                &  29.44             &       91.85             & 344.3                       &  11.05                 & 79.08                  & 22.61                                     \\ \hline
Low Rank   Based: PSTNN\cite{13-PSTNN}                              &   14.85                 &     66.13              &       44.17               & 22.40                   & 77.95                   & 29.11                                       \\ \hline
Low Rank   Based: MSLSTIPT \cite{3-anti-miss}                       &   8.342                 &    47.40              &  888.1                & 10.30                    & 82.13               &  1131                                     \\ \hline
CNN Based: MDvsFA-cGAN  \cite{24-ICCV19}                              &  75.14                 &     90.47              &   25.34                  &  60.30                 &    89.35              & 56.35                                     \\ \hline
CNN Based:   ACM  \cite{22-ACM}                                       &  67.08                   &  95.97                &   10.18                &   70.33                   &   93.91               &   3.728                                           \\ \hline
CNN Based:   ALCNet  \cite{23-ALCNet}                                 &  81.40                  &  96.51               &  9.261                   &   73.33                  &    96.57               &     30.47                                         \\ \hline
\textbf{DNA-Net-VGG10   (ours)}                             & 85.23                    & 96.95                  & 6.782                   & 74.96                     & 97.34                   & 26.73                                            \\ \hline
\textbf{DNA-Net-ResNet10   (ours)}                          & 86.36                    & 97.39                  & 6.897                  & 76.24                     & 97.71                   & 12.80                                       \\ \hline
\textbf{DNA-Net-ResNet18   (ours)}                          & \textcolor{red} {87.09}         & \textcolor{red} {98.73}        &\textcolor{blue} {4.223}         & \textcolor{blue} {77.47}             & \textcolor{red} {98.48}        & \textcolor{red} {2.353}                                          \\ \hline
\textbf{DNA-Net-ResNet34   (ours)}                          & \textcolor{blue} {86.87}        & \textcolor{blue} {97.98}       & \textcolor{red} {3.710}          & \textcolor{red} {77.54}             & \textcolor{blue} {98.10}           & \textcolor{blue} {2.510}                                          \\ \hline
\end{tabular}
\end{table*}

\subsubsection{Intersection over Union}
Intersection over Union ($IoU$) is a pixel-level evaluation metric. It evaluates profile description ability of the algorithm. IoU is calculated by the ratio of intersection and the union areas between the predictions and labels, i.e.,
 {\begin{equation}\label{mIoU}
{IoU}= \frac{A_{inter}}{A_{Union}},
  \end{equation}where $A_{inter}$ and $A_{Union}$ represent the interaction areas and union areas, respectively.}

\subsubsection{Probability of Detection}
Probability of Detection (${P}_{d}$) is a target-level evaluation metric. It measures the ratio of correctly predicted target number $T_{correct}$ over all target number $T_{All}$.  ${P}_{d}$ is defined as follows:

 {\begin{equation}\label{PD}
  {P}_{d}= \frac{T_{correct}}{T_{All}}.
  \end{equation}}

  If the centroid deviation of the target is less than the pre-defined deviation threshold $ D_{\textit{thresh}}$, we consider those targets as correctly predicted ones. We set the pre-defined deviation threshold as 3 in this paper.

\subsubsection{False-Alarm Rate}
False-Alarm Rate (${F}_{a}$) is another target-level evaluation metric. It is used to measure the ratio of falsely predicted pixels $P_{false}$ over all image pixels $P_{All}$. ${F}_{a}$ is defined as follows:

 {\begin{equation}\label{FA}
 {F}_{a} = \frac{P_{false}}{P_{All}}.
  \end{equation}}

If the centroid deviation of the target is larger than the pre-defined deviation threshold, we consider those pixels as falsely predicted ones. We set the pre-defined deviation threshold as 3 in this paper.

\subsubsection{Receiver Operation Characteristics}

Receiver Operation Characteristics (ROC) is used to describe the changing trends of the detection probability (${P}_{d}$) under varying false alarm rate (${F}_{a}$).

\begin{figure*}
 \centering
 \includegraphics[width=18.2cm]{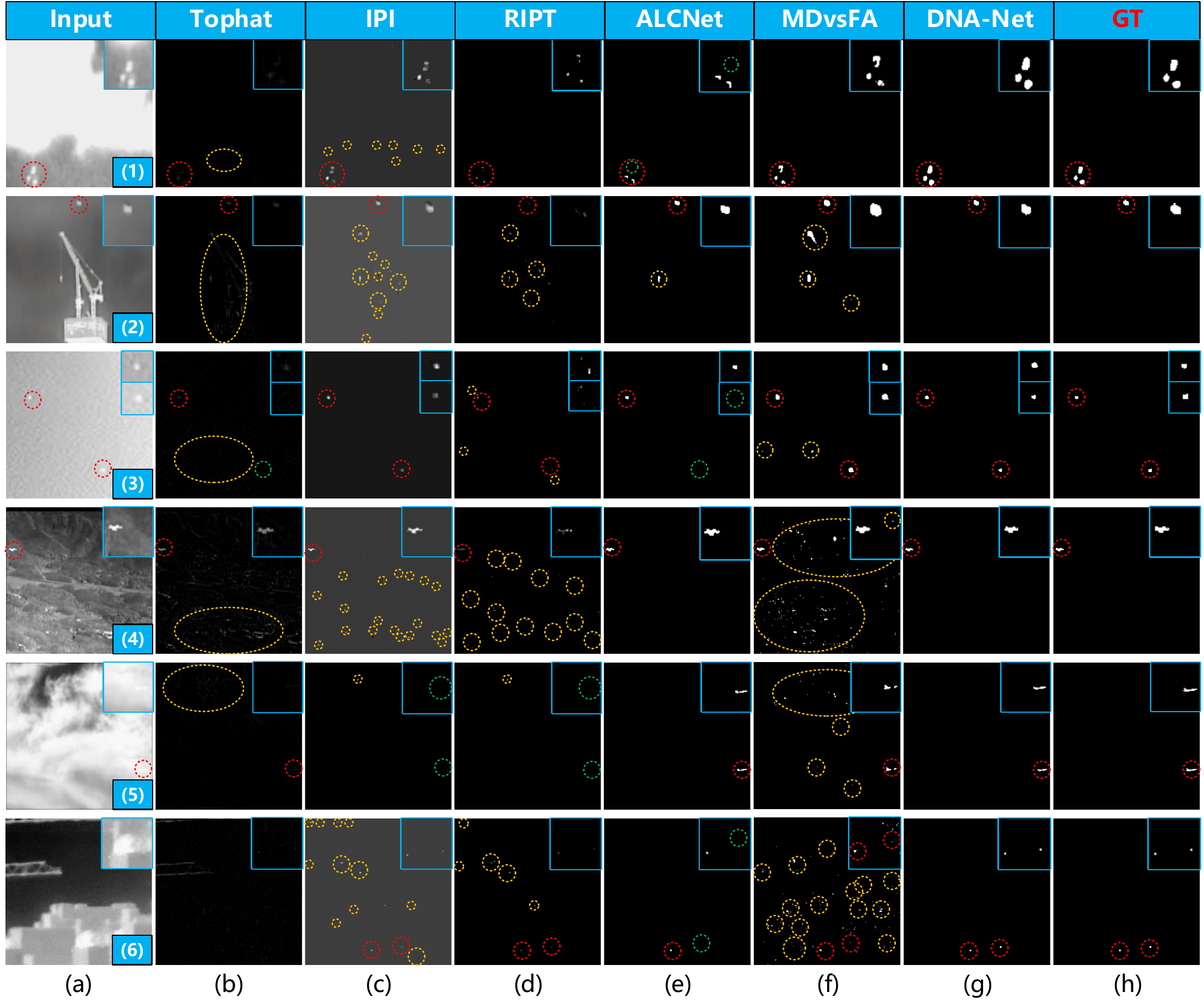}
 \caption{Qualitative results achieved by different SIRST detection methods. For better visualization, the target area is enlarged in the right-top corner. The correctly detected target, false alarm, and miss detection areas are highlighted by red, yellow, and green dotted circles, respectively. Our DNA-Net can generate output with precise target localization and shape segmentation under a lower false alarm rate.}\label{SOAT_1}
 \end{figure*}

\subsection{Implementation Details}\label{Protocol}

As discussed in Section \ref{SOAT}, we used the published NUAA-SIRST dataset\cite{23-ALCNet} and our NUDT-SIRST dataset for both training and test. Previous works \cite{23-ALCNet,22-ACM} set the train-to-test ratios as 3 (i.e., 256 images for training and 86 images for testing). However, sufficient test images are crucial to evaluate the real performance of the model. Therefore, we set the train-to-test ratio to 1 (i.e., 213 images for training and 214 images for testing). Before training, all input images were first normalized. Then, these normalized images were sequentially processed by random image flip, blurring, and crop for data augmentation. Next, these images were resized to a resolution of 256 $\times$ 256 before being fed into the network.

In this paper, we adopted a segmentation network as our baseline to generate a pixel-level segmentation map and then used a clustering algorithm to achieve target localization. The U-net paradigm with ResNets\cite{30_resnet} was chosen as our segmentation backbone. The number of down-sampling layer $i$ was chosen as 4. Our network was trained using the Soft-IoU loss function and optimized by the Adagrad method\cite{27-Adagrad} with the CosineAnnealingLR scheduler. We initialized the weights and bias of our model using the Xavier method\cite{28-xavier}. We set the learning rate, batch size, and epoch size as 0.05, 16, and 1500, respectively. All models were implemented in PyTorch \cite{29-pytorch} on a computer with an AMD Ryzen 9 3950X @ 2.20 GHz CPU and an Nvidia GeForce 3090 GPU.

\begin{figure*}
 \centering
 \includegraphics[width=18.2cm]{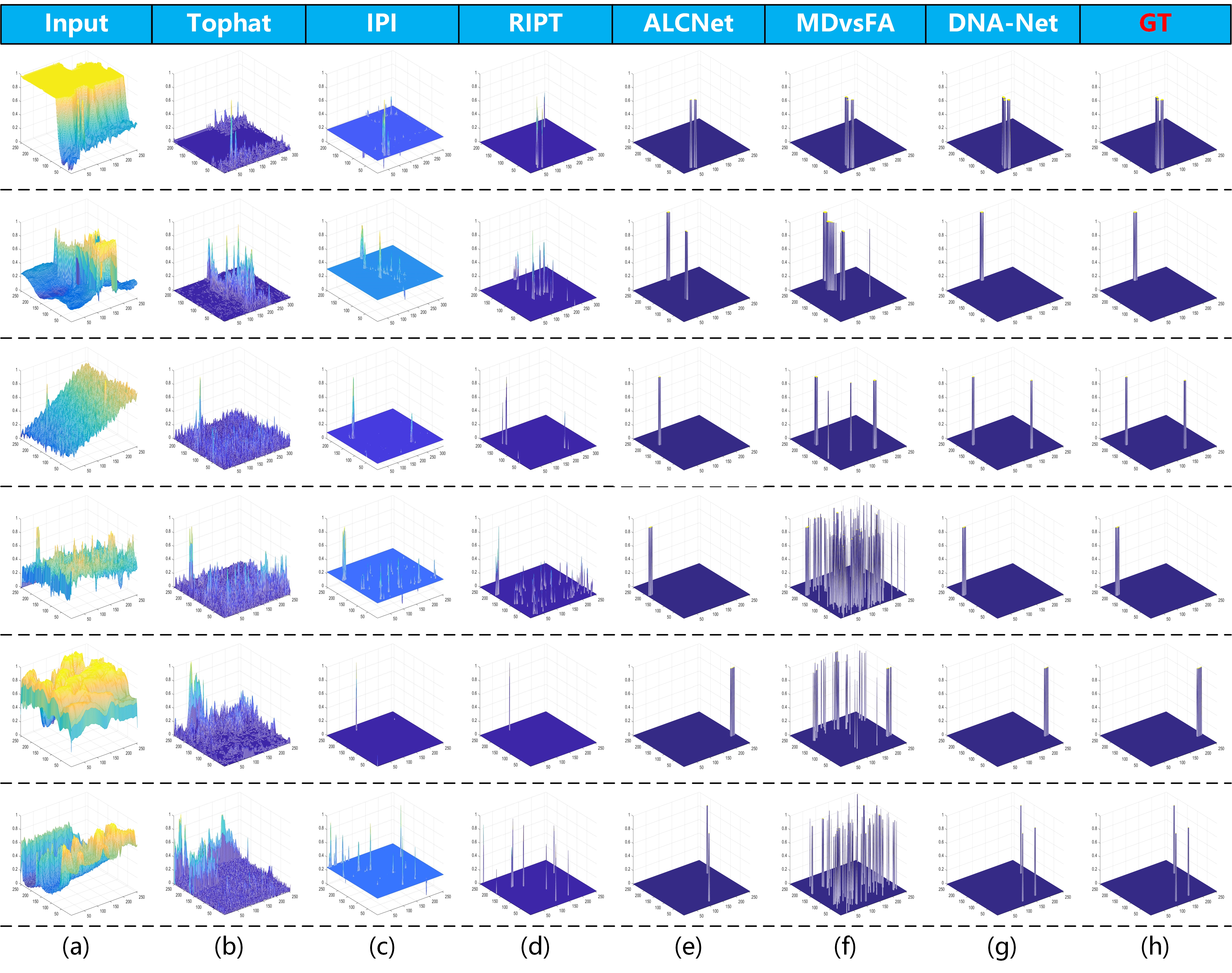}
 \caption{3D visualization results of different methods on 6 test images. }\label{SOAT_2}
 \end{figure*}

\subsection{Comparison to the State-of-the-art Methods}\label{SOAT}

 To demonstrate the superiority of our method, we compare our DNA-Net to several state-of-the-art (SOTA) methods, including traditional methods (Top-Hat\cite{4-tophat}, Max-Median \cite{5-maxmedian}, WSLCM \cite{9-WSLLCM}, TLLCM\cite{8-TLLCM}, IPI\cite{10-IPI}, NRAM\cite{11-NRAM}, RIPT\cite{12-RIPT}, PSTNN\cite{13-PSTNN}, MSLSTIPT \cite{3-anti-miss}) and CNN-based methods (MDvsFA-cGAN\cite{24-ICCV19}, ACM\cite{22-ACM}, ALCNet\cite{23-ALCNet}) on the NUAA-SIRST and NUDT-SIRST datasets \footnote{Note that, we follow ACM \cite{22-ACM} and ALCNet \cite{23-ALCNet} to not use the NUST-SIRST for comparison in the main body of our manuscript since only about 30\% of targets meet the SPIE's definition of small targets. To achieve a more comprehensive comparison, we have updated the experimental
results of NUST-SIRST and released the trained model at our Github repository.}. For fair comparison, we retrained all the CNN-based methods on the same training datasets as our DNA-Net. It is worth noting that we use our implementations for these methods for fair comparison. Most of these open-source CNN-based codes are rewritten by pytorch and released at: \url{https://github.com/YeRen123455/Infrared-Small-Target-Detection}.

\subsubsection{Quantitative Results}

For all the compared algorithms, we first obtained their predicts and then performed noise suppression by setting a threshold to remove low-response areas. Specifically, the adaptive threshold (${T}_{adaptive}$) was calculated for traditional methods according to:

\begin{equation}\label{threshold}
\emph{T}_{adaptive}=Max[Max(\mathbf{G})\times0.7,0.5\times\sigma(\mathbf{G})+avg(\mathbf{G})],
\end{equation} where $Max(\mathbf{G})$ represents the largest value of output. ${T}_{adaptive}$ represents adaptive threshold. $\sigma(\mathbf{G})$ and $avg(\mathbf{G})$ mean the standard derivation and average value of output, respectively. For CNN-based methods, we followed their original papers and adopted their fixed thresholds (i.e., 0, 0, 0.5 for ACM\cite{22-ACM}, ALCNet\cite{23-ALCNet}, and MDvsFA-cGAN\cite{24-ICCV19}, respectively). We kept all remaining parameters the same as their original papers.

\arrayrulewidth=0.75pt
\begin{table}[h]\scriptsize
\centering
\captionsetup[table]{labelformat=simple, labelsep=newline, justification=centering, textfont=sc}
\renewcommand\arraystretch{1.1}
\caption{Comparision to SOTA methods in terms of train time, inference time, and $IoU$($\times10^{2}$)$/$$P_{d}$($\times10^{2}$)$/$$F_{a}$($\times10^{6}$) on the NUDT-SIRST dataset.} \label{Tab_compute_eff}
\label{DiagnosisResults}
\begin{tabular}{|c|c|c|c|}
\hline
\multirow{2}{*}{\begin{tabular}[c]{@{}c@{}}Method  \end{tabular}} & \multicolumn{3}{c|}{Evaluation Metircs}                 \\ \cline{2-4}
                                                         &   Train Time
                                                         &   Inference Time
                                                         & $IoU$/$P_{d}$/$F_{a}$   \\ \hline
MDvsFA-cGAN              \cite{24-ICCV19}      & 9.952h              & 0.019s             & 75.14/90.47/25.34            \\ \hline
ACM   (ResNet20)         \cite{22-ACM}         &  \textbf{0.946}h    &  \textbf{0.011}s   & 67.08/95.97/10.18           \\ \hline
ALCNet   (ResNet20)     \cite{23-ALCNet}       &  7.623h             & 0.021s              & 81.40/96.51/9.261           \\ \hline
\textbf{DNA-Net-ResNet10-Light}                  &  3.862h            & 0.012s              &   \textbf{83.68/97.25/13.23}         \\ \hline
\end{tabular}
\end{table}

Quantitative results are shown in Table~\ref{TabOVER_ALL}. The improvements achieved by our DNA-Net over traditional methods are significant. That is because, both NUDT-SIRST and NUAA-SIRST contain challenging images with different SCR, clutter background, target shape, and target size. Our DNA-Net can learn discriminative features robust to scene variations. In contrast, the traditional methods are usually designed for specific scenes (e.g., specific target size and clutter background). The manually-selected parameters (e.g., structure size in Tophat and patch size in IPI) limit the generalization performance of these methods. Moreover, we also observe that the IoU improvements are obviously higher than the improvement of ${P}_{d}$ and ${F}_{a}$. That is because, the traditional methods mainly focus on the overall localization of the target instead of precise shape matching. It also validates our claim that using pixel-level evaluation metric (such as $IoU$) introduces unfair comparison and leads to inaccurate conclusion.

As shown in Table~\ref{TabOVER_ALL}, the improvements achieved by DNA-Net over other CNN-based methods (i.e., MDvsFA-cGAN, ACM, and ALCNet) are obvious. That is because, we redesign a new backbone network that is tailored for SIRST detection.
\arrayrulewidth=0.75pt
\begin{table}[h]\scriptsize
\centering
\captionsetup[table]{labelformat=simple, labelsep=newline, justification=centering, textfont=sc}
\renewcommand\arraystretch{1.1}
\caption{$P_{d}$($\times10^{2}$)$/$$F_{a}$($\times10^{6}$) values achieved by different state-of-the-art methods on the NUDT-SIRST dataset with different settings of $D_{\textit{thresh}}$. } \label{Tab2_derivation}
\label{DiagnosisResults}
\begin{tabular}{|c|c|c|c|}
\hline
\multirow{2}{*}{\begin{tabular}[c]{@{}c@{}}Method  \end{tabular}} & \multicolumn{3}{c|}{Maximum Centroid Deviation }                 \\ \cline{2-4}
                                                         &$D_{\textit{thresh}}$\textless 2
                                                         & $D_{\textit{thresh}}$\textless 3
                                                         &$D_{\textit{thresh}}$\textless 4  \\ \hline
MDvsFA-cGAN              \cite{24-ICCV19}      & 89.31/30.15      & 90.47/25.34   & 91.21/24.98           \\ \hline
ACM   (ResNet20)         \cite{22-ACM}         &  95.56/15.65    &  95.97/10.18   & 95.97/10.18          \\ \hline
ALCNet   (ResNet20)     \cite{23-ALCNet}       & 96.30/10.36    &   96.51/9.262   & 96.73/9.581           \\ \hline
\textbf{DNA-Net-ResNet18}                                &  \textbf{98.51/4.987}     & \textbf{98.73/4.228}   &  \textbf{98.73/4.228}        \\ \hline
\end{tabular}
\end{table}
The U-shape basic backbone with our dense nested interactive skip connection module can achieve progressive feature fusion and selectively enhance the informative features in deep CNN layers. Consequently, intrinsic features of infrared small targets can be maintained and fully learned in the network. It is also worth noting that the $IoU$ improvements of our method on NUDT-SIRST is significantly higher than those on the NUAA-SIRST dataset. That is because, our dataset contains more challenging scenes with various target sizes, types and poses. Our channel and spatial attention module and feature pyramid fusion module help to learn discriminative features to achieve better performance.

Quantitative results in Table~\ref{Tab2_derivation} demonstrate that our method is superior to other deep-learning based methods under different pre-defined deviation thersholds.

\subsubsection{Qualitative Results}

\begin{figure*}[htbp]
\centering
\subfigure[]{
\begin{minipage}[t]{0.33\linewidth}
\centering
\includegraphics[width=6.0 cm]{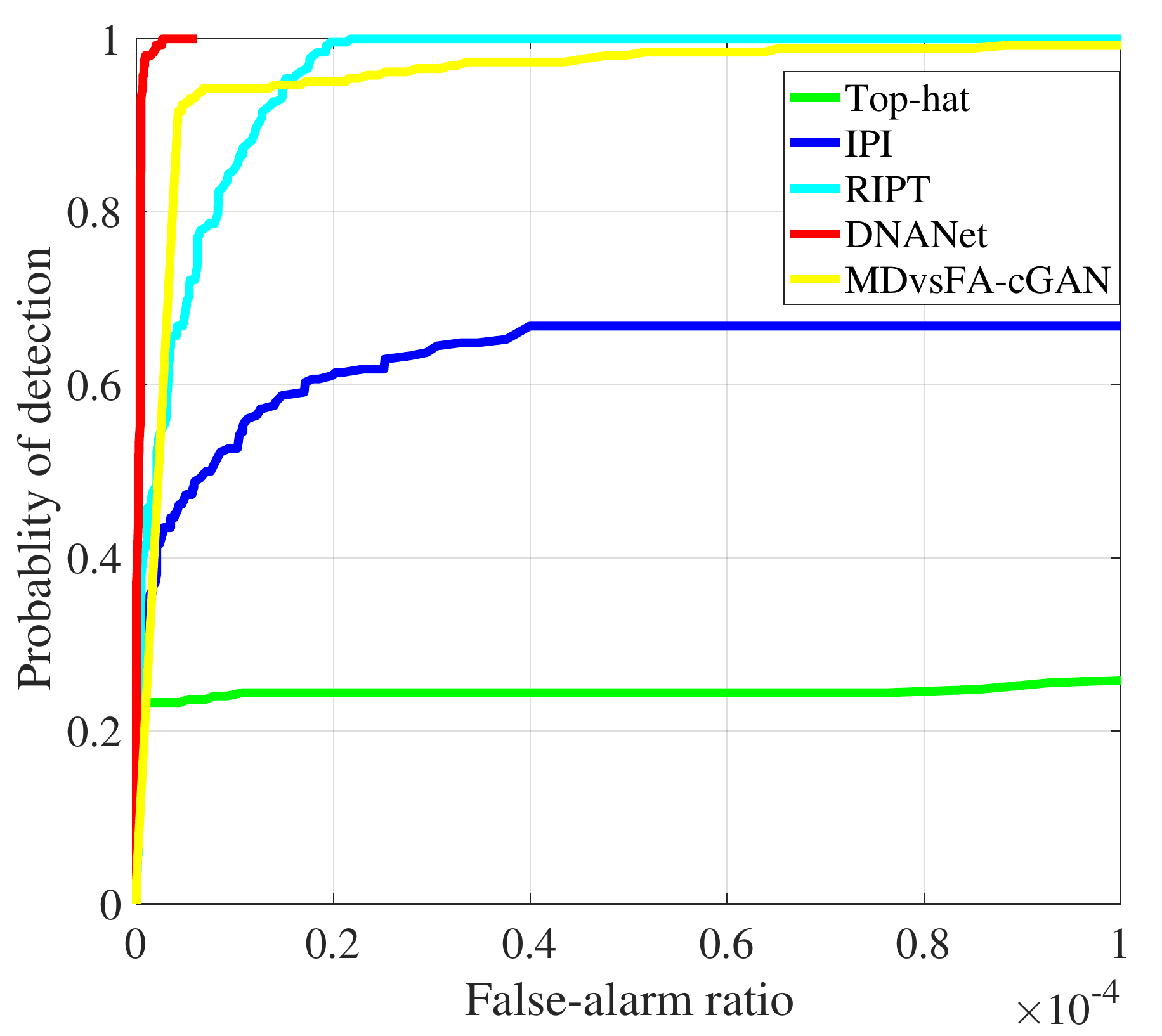}
\end{minipage}%
}%
\subfigure[]{
\begin{minipage}[t]{0.33\linewidth}
\centering
\includegraphics[width=6.0 cm]{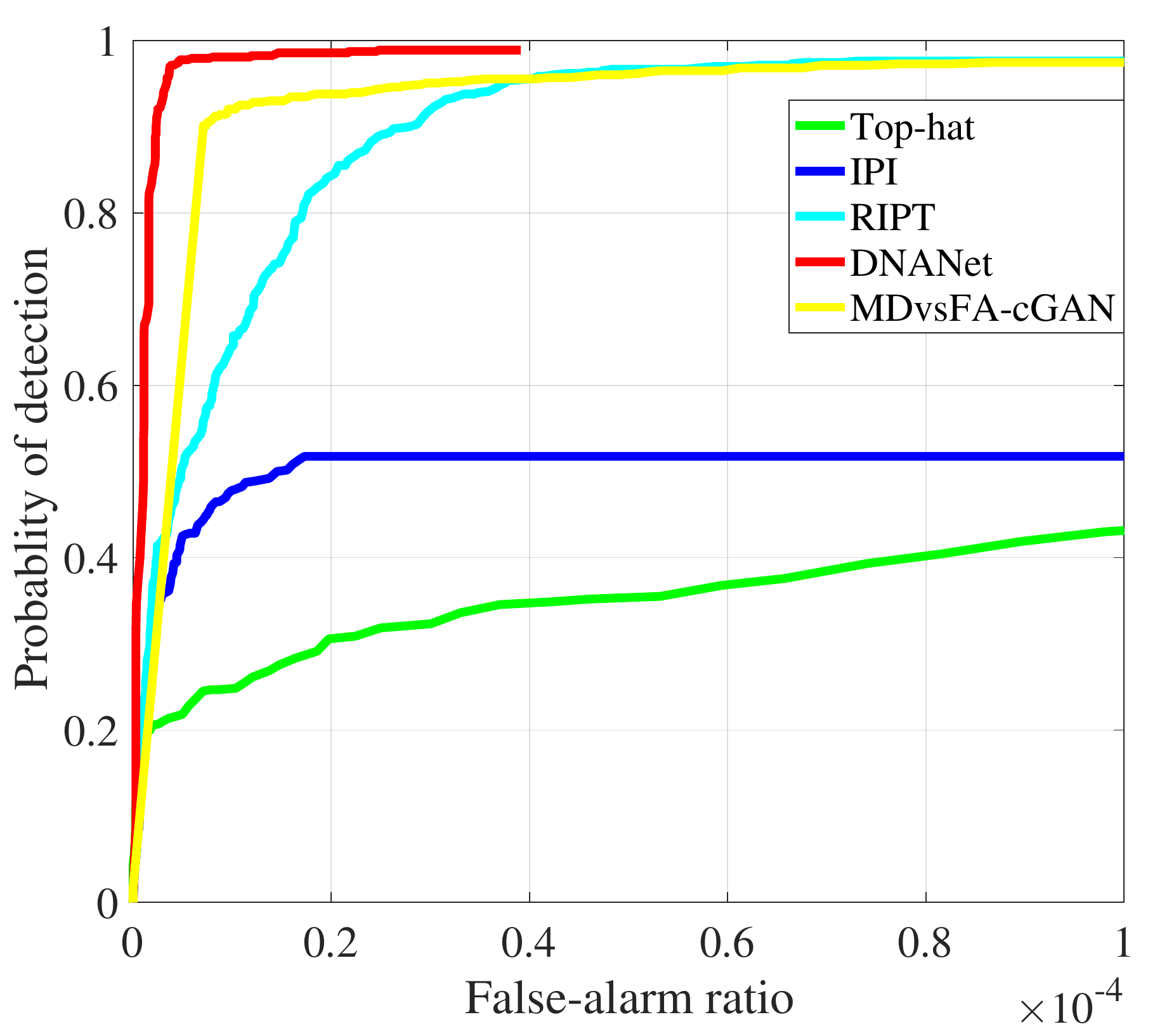}
\end{minipage}%
}%
\subfigure[]{
\begin{minipage}[t]{0.33\linewidth}
\centering
\includegraphics[width=6.0 cm]{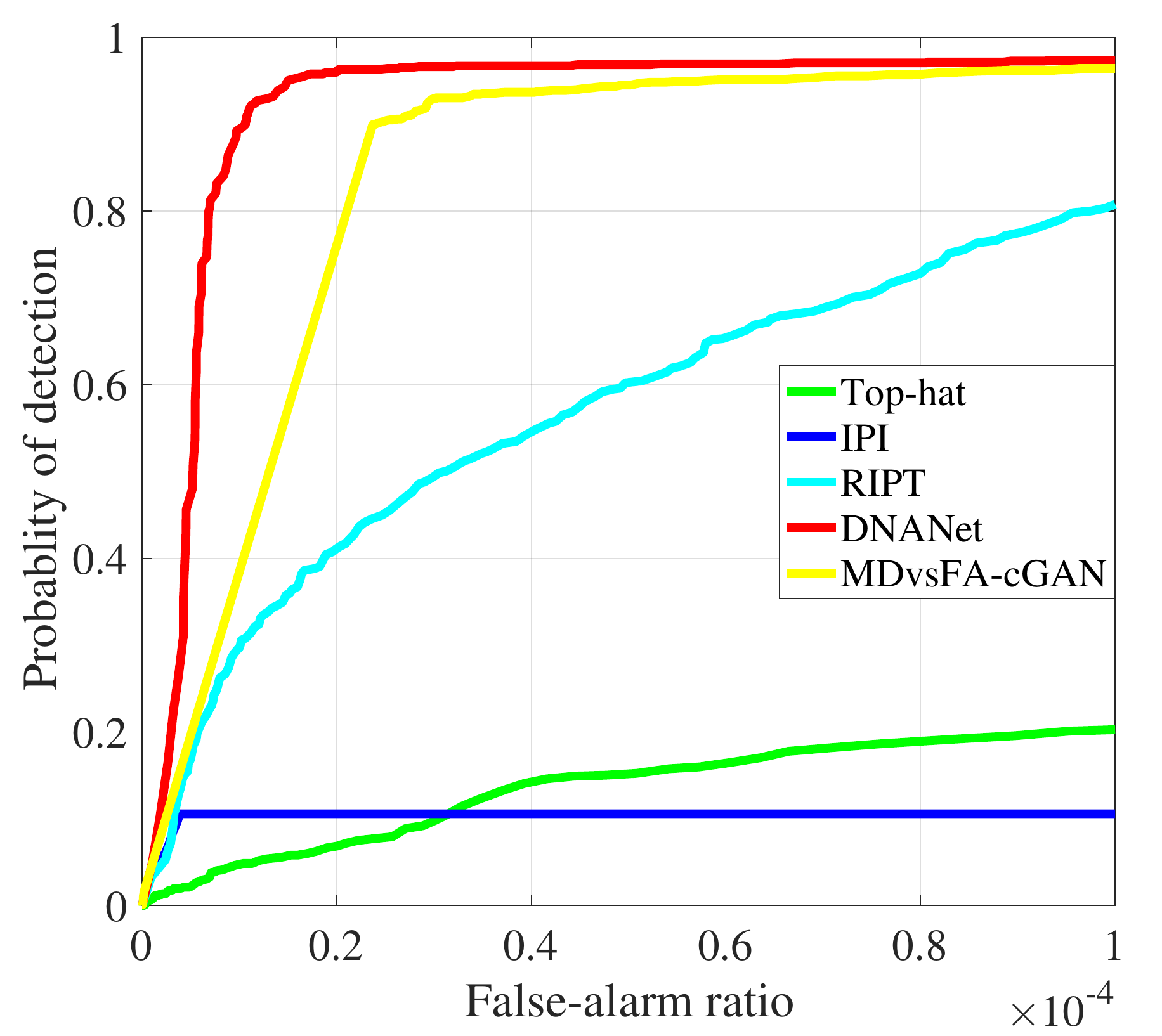}
\end{minipage}%
}%
\caption{ROC performance on (a) point targets subset, (b) point targets subset + spot targets subset, (c) all kinds of targets of NUDT-SIRST. With the increase of spot and extended targets ratio, the performance of traditional methods suffers dramatic drop. In contrast, the performance our DNA-Net is stable.}\label{ROC}
\end{figure*}

Qualitative results on two datasets (i.e., NUDT-SIRST, NUAA-SIRST) are shown in Fig.~\ref{SOAT_1} and Fig.~\ref{SOAT_2}. Compared with traditional methods, our method can produce output with precise target localization and shape segmentation under very low false alarm rate. Nonetheless, the traditional methods only perform well on point targets, (e.g., image-3), and easily generate lots of false alarm areas in local highlight areas (e.g., image-4 and image-6). Moreover, as shown in Fig.~\ref{ROC}, we divided our NUDT-SIRST dataset into point targets subset, spot targets subset, and extended targets subset. With the increase of spot and extended targets ratio, traditional methods suffers dramatic performance decrease while our DNA-Net maintains high accuracy. That is because, the performance of traditional methods rely heavily on handcrafeted features and cannot adapt to the variations of target sizes.

The CNN-based methods (i.e., MDvsFA-cGAN, ACM, and ALCNet) perform much better than traditional methods. However, due to the complicated scenes in our NUDT-SIRST, MDvsFA-cGAN produces many false alarm and miss detection areas (Fig.~\ref{SOAT_2}). Our DNA-Net is more robust to these scene changes.
Moreover, our DNA-Net can generate better shape segmentation than ALCNet. That is because, our designed new backbone can well adapt to various clutter background, target shape, and target size challenges and thus achieves better performance.

\subsubsection{Computational Efficiency}

In this part, we reduced half of the channels in DNA-Net-ResNet10 to build DNA-Net-ResNet10-Light and compared it to several competitive methods (i.e., MDvsFA-cGAN\cite{24-ICCV19}, ACM\cite{22-ACM}, ALCNet\cite{23-ALCNet}) in terms of training time and inference time. As shown in Table~\ref{Tab_compute_eff}, our DNA-Net-ResNet10-Light achieves the highest $IoU$, $P_{d}$, and the lowest $F_{a}$ with comparable training and inference time. This clearly demonstrates the high computational efficiency of our method.

\subsection{Ablation Study}\label{SOAT}

In this subsection, we compare our DNA-Net with several variants to investigate the potential benefits introduced by our network modules and design choice.

\subsubsection{The Dense Nested Interactive Module (DNIM)}
The dense nested interactive skip-connection module is used to interact with features at different scale levels to enlarge receptive fields while maintain fine-grained features at the finest scale level.
To demonstrate the effectiveness of our DNIM, we introduced three network variants and made   their model sizes comparable for fair comparison.

\begin{table}[]\scriptsize
\centering
\renewcommand\arraystretch{1.1}
\caption{$IoU$($\times10^{2}$)$/$$P_{d}$($\times10^{2}$)$/$$F_{a}$($\times10^{6}$) values achieved by main variants of DNA-Net and DNIM on the NUDT-SIRST and NUAA-SIRST datasets. Top-to-bottom and Left-to-right mean stack U-shape sub-network from different directions} \label{Tab_DNIM}
\begin{tabular}{|c|c|c|c|}
\hline
\multirow{2}{*}{Model} & \multirow{2}{*}{\#Params(M)} & \multicolumn{2}{c|}{Datasets}                   \\ \cline{3-4}
                       &                           & NUDT-SIRST    & NUAA-SIRST                      \\ \hline
DNA-Net w/o DNIM       &        4.71                        & 85.01/96.50/8.521               & 75.12/97.34/12.05             \\ \hline
DNA-Net-top-to-bottom       &   4.72                        & 85.75/96.96/7.682                & 75.94/97.71/11.84             \\ \hline
DNA-Net-left-to-right       &   4.71                        & 85.89/97.29/4.649               & 76.59/98.10/11.05             \\ \hline
DNA-Net-ResNet18       &     4.70                          & 87.09/98.73/4.223               & 77.47/98.48/2.353              \\ \hline
\end{tabular}
\end{table}

\begin{figure*}
\centering
\includegraphics[width=18.2cm]{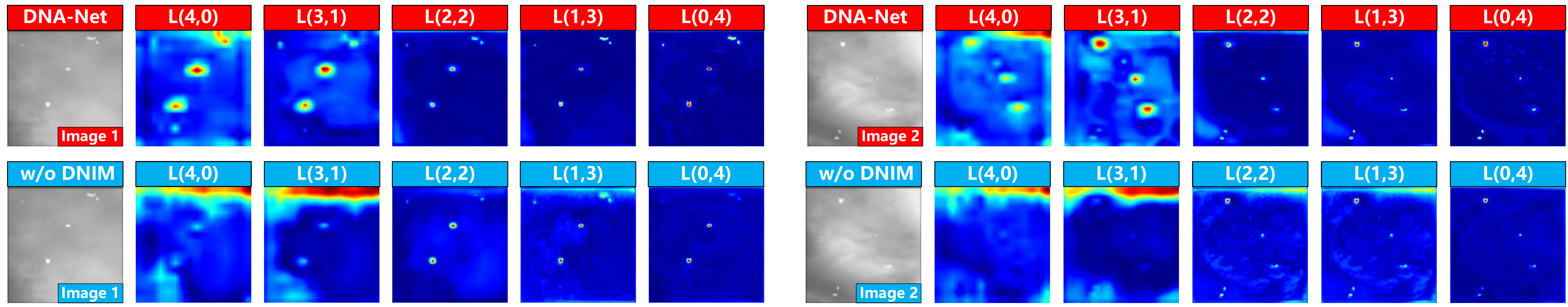}
\caption{Visualization map of DNA-Net (row 1) and DNA-Net w/o DNIM  (row 2). The feature maps from the deep layer of DNA-Net w/o DNIM loses representation of small targets. It finally results in low values and miss detection in the output layer.}\label{Compare_DNIM}
\end{figure*}

Table~\ref{Tab_DNIM} shows the comparative results achieved by \textit{DNA-Net} and its variants. It can be observed that the $IoU$, ${P}_{d}$, and ${F}_{a}$ values of \textit{DNA-Net w/o DNIM} suffer decreases of 2.08\%, 2.23\%, and an increase of 4.298$\times 10^{-6}$ on the NUDT-SIRST dataset. Similar results are also observed on the NUAA-SIRST dataset. That is because, DNIM progressively aggregates features at multiple scales to maintain the target information at the finest scale for better performance. Visualization maps shown in Fig.~\ref{Compare_DNIM} also demonstrates the effectiveness of our DNIM. Small targets are lost in the feature maps of the deep layer in DNA-Net w/o DNIM (i.e., L(4,0), L(3,1)).

\begin{figure}
\centering
\includegraphics[width=8.8cm]{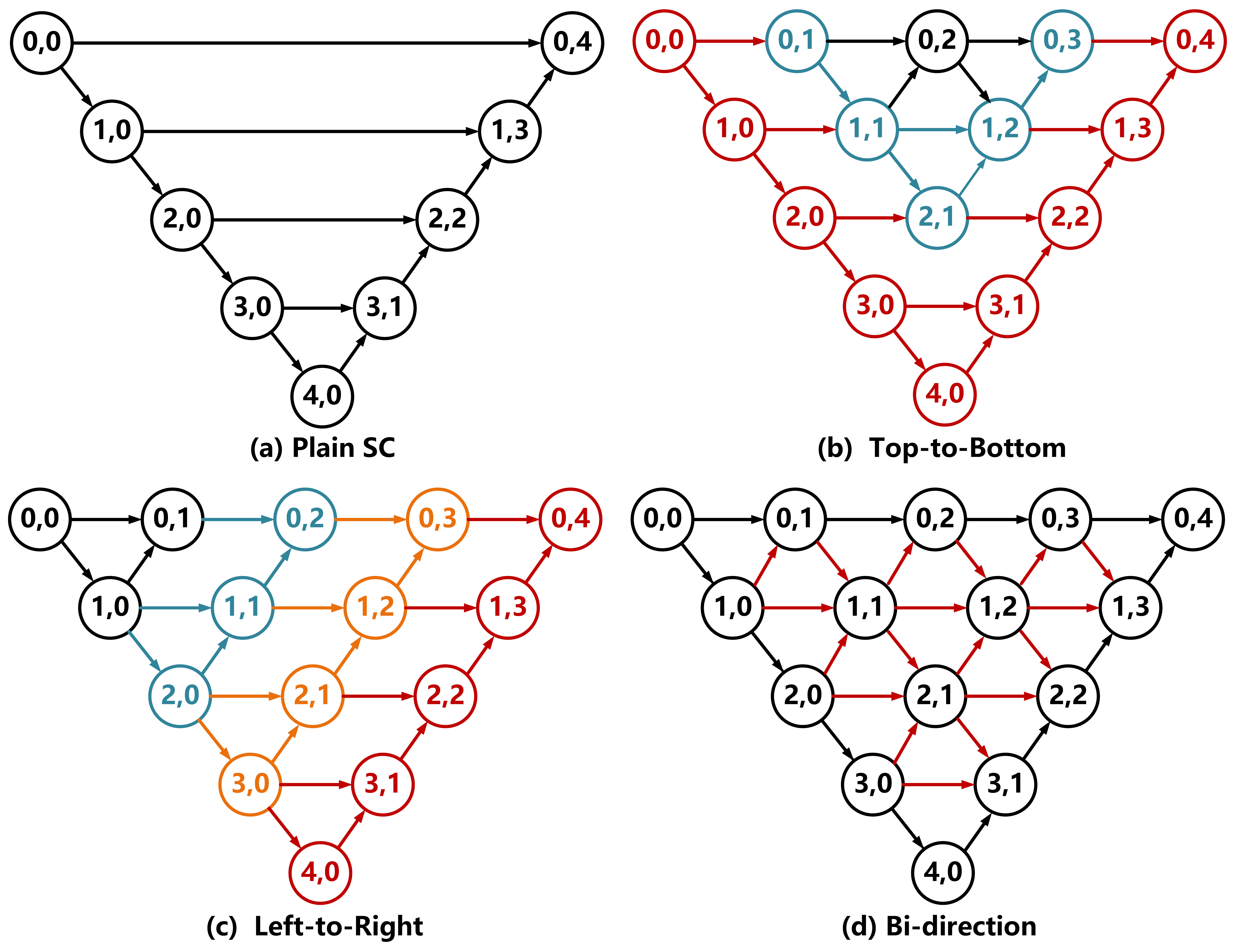}
\caption{Three variants of DNIM. (a) DNA-Net w/o DNIM. (b) DNA-Net-top-to-bottom. (c) DNA-Net-left-to-right. (d) DNA-Net, each color represents different U-shape sub-networks.}\label{Ablation_DNIM}
\end{figure}

\begin{itemize}
\item \textbf{DNA-Net w/o DNIM}: We replaced the dense nested interactive skip connection module with a regular plain skip connection module.
\item \textbf{DNA-Net-left-to-right}: As shown in Fig.~\ref{Ablation_DNIM}(c), multiple U-shape subnetworks with different depths are stacked from left to right. Each node in the middle part of the network can receive features from its own and the lower layer.
\item \textbf{DNA-Net-top-to-bottom}: We stacked the U-shape subnetworks from top to bottom to generate \textit{DNA-Net-top-to-bottom}, as shown in Fig.~\ref{Ablation_DNIM}(b). Different from \textit{DNA-Net-left-to-right}, this variant stacks U-shape subnetworks with three kinds of depth and only its core part uses tri-direction skip connection.
\end{itemize}

As shown in Table~\ref{Tab_DNIM}, \textit{DNA-Net-left-to-right} suffers decreases of 1.20\%, 1.44\%, and an increase of 0.426 $\times 10^{-6}$ in terms of $IoU$, ${P}_{d}$, and ${F}_{a}$ values over \textit{DNA-Net} on the NUDT-SIRST dataset. That is because, each node in \textit{DNA-Net-left-to-right} only interacts with the deep layer instead of full interaction among shallow, their-own, and deep layers. Shallow layer has rich localization and profile information, but the information is not fully incorporated at the skip connection stage. Consequently, this variant has limited performance.

As compared to our DNA-Net, the variant \textit{DNA-Net-top-to-bottom} suffers decreases of 1.34\%, 1.77\%, and an increase of 3.459 $\times 10^{-6}$ in terms of $IoU$, ${P}_{d}$, and ${F}_{a}$ values on NUDT-SIRST dataset. That is because, only the core part of this variant adopts tri-direction skip connection, the remaining part still uses the plain skip connection. Moreover, its tri-direction interactive area is relatively shallow, high-level information can not be fully exploited at shallow layers.

\subsubsection{The Channel and Spatial Attention Module (CSAM)}
The channel and spatial attention module is used for adaptive feature enhancement to achieve better feature fusion. To investigate the benefits introduced by this module, we compare our DNA-Net with four variants. To achieve fair comparison (i.e., comparable model size), we increased the number of filters of all convolution layers of four variants to make their model sizes slightly larger than \textit{DNA-Net}.

\begin{table}[]\scriptsize
\centering
\renewcommand\arraystretch{1.1}
\caption{$IoU$($\times10^{2}$)$/$$P_{d}$($\times10^{2}$)$/$$F_{a}$($\times10^{6}$) values achieved by main variants of DNA-Net and CSAM on the NUDT-SIRST and NUAA-SIRST datasets. $\oplus$ means element-wise summing as feature fusion method.} \label{Tab_CSAM}
\begin{tabular}{|p{2.4cm}<{\centering}|c|c|c|}
\hline
\multirow{2}{*}{Model} & \multirow{2}{*}{\#Params(M)} & \multicolumn{2}{c|}{Datasets}                   \\ \cline{3-4}
                       &                           & NUDT-SIRST    & NUAA-SIRST                      \\ \hline
DNA-Net w/o CSAM       &       4.70                  & 85.90/96.62/5.738              & 75.81/96.19/22.12               \\ \hline
DNA-Net w/o CSAM$\oplus$       &  4.71               & 85.25/96.62/6.710               & 75.35/95.82/34.97             \\ \hline
DNA-Net w/o CA       &     4.73                      & 86.27/96.96/4.881             & 76.20/96.96/12.69                \\ \hline
DNA-Net w/o SA       &      4.73                     & 86.14/96.73/4.128             & 76.69/97.34/10.96                \\ \hline
DNA-Net-ResNet18       &    4.70                    & 87.09/98.73/4.223               & 77.47/98.48/2.353                \\ \hline
\end{tabular}
\end{table}

\begin{figure*}
\centering
\includegraphics[width=18.2cm]{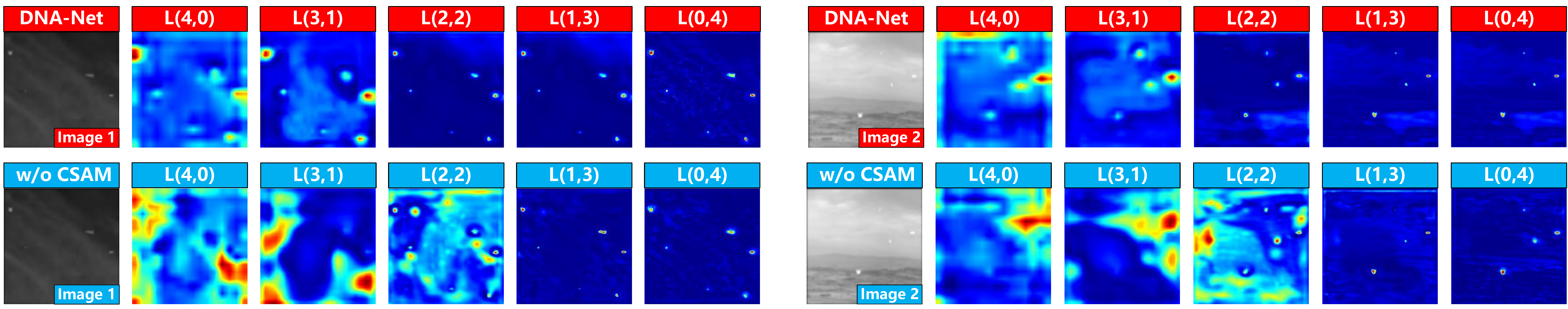}
\caption{Visualization map of DNA-Net (row 1) and DNA-Net w/o CSAM (row 2). The feature maps from the deep layer of DNA-Net have high values representation to informative cues and finally results in precise profile segmentation in output layer.}\label{Compare_CS}
\end{figure*}

\begin{itemize}
\item \textbf{DNA-Net w/o CSAM}: We removed the channel and spatial attention module in this variant and directly concatenate multi-layer features for subsequent process.
\item \textbf{DNA-Net w/o CSAM (Element-wise summation)}: We replaced CSAM with common element-wise summation in this variant to explore the effectiveness of CSAM. Specifically, we used 1$\times$1 convolution operation and up-sampling/down-sampling to make features from different layer identical. Then, an element-wise summation is used to achieve multi-layer feature fusion.
\item \textbf{DNA-Net w/o channel attention}: We removed the channel attention operation in this variant to evaluate its contribution.
\item \textbf{DNA-Net w/o spatial attention}: We canceled the spatial attention operation in this variant to investigate the benefit introduced by spatial attention.
\end{itemize}

If CSAM is removed, the performance suffers decreases of 1.19\%$/$1.84\%, 2.11\%$/$2.11\%, and an increase of 1.515$/$2.487 $\times 10^{-6}$ in terms of $IoU$, ${P}_{d}$, and ${F}_{a}$ for \textit{DNA-Net w/o CSAM} and \textit{DNA-Net w/o CSAM $\oplus$} on the NUDT-SIRST dataset, respectively. Similar results are achieved on the NUAA-SIRST dataset. This clearly demonstrates the importance of the channel and spatial attention module. As shown in Fig.~\ref{Compare_CS}, with the help of CSAM, the feature maps from the deep layer of DNA-Net have high response to informative cues and finally results in precise shape segmentation.

\begin{table}[]\scriptsize
\centering
\renewcommand\arraystretch{1.1}
\caption{$IoU$($\times10^{2}$)$/$$P_{d}$($\times10^{2}$)$/$$F_{a}$($\times10^{6}$) values achieved by main variants of DNA-Net and FPFM on the NUDT-SIRST and NUAA-SIRST datasets. DNA-Net w/o $L_{i, j , k}$ means the outputs from layer $i$, $j$, and $k$ are removed from FPFM.} \label{Tab_FPFM}
\begin{tabular}{|p{2.4cm}<{\centering}|c|c|c|}
\hline
\multirow{2}{*}{Model} & \multirow{2}{*}{\#Params(M)} & \multicolumn{2}{c|}{Datasets}                   \\ \cline{3-4}
                       &                           & NUDT-SIRST    & NUAA-SIRST                      \\ \hline
DNA-Net w/o FPFM       &  4.72          &  86.26/96.84/6.033    & 76.87/97.71/12.97               \\ \hline
DNA-Net w/o L345       &  4.70          &  86.38/96.38/5.287    & 76.34/97.34/12.83          \\ \hline
DNA-Net w/o L45        &  4.70          &  86.29/97.13/6.314    & 76.92/98.10/6.913                \\ \hline
DNA-Net w/o L5         &  4.70          &  86.86/97.89/7.236    & 77.11/98.10/6.342              \\ \hline
DNA-Net-ResNet18       &  4.70          &  87.09/98.73/4.223    & 77.47/98.48/2.353                \\ \hline
\end{tabular}
\end{table}

As shown in Table~\ref{Tab_CSAM}, \textit{DNA-Net w/o channel attention} suffers decreases of 0.82\%, 1.77\%, and an increase of 0.658 $\times 10^{-6}$ in terms of $IoU$, ${P}_{d}$, and ${F}_{a}$ values over \textit{DNA-Net} on NUDT-SIRST dataset. That is because, channel attention unit in our DNA-Net can better exploit informative channels to enhance the representation capability of features.

If the spatial attention unit is removed, the performance suffers decreases of 0.95\%, 2.00\%, and an increase of 0.095 $\times 10^{-6}$ in terms of $IoU$, ${P}_{d}$, and ${F}_{a}$ values  for \textit{DNA-Net} on NUDT-SIRST dataset. That is because, infrared small targets are easily immersed in heavy cloud and noise, it is hard to distinguish these small and dim targets from the background. Spatial attention facilitates the network to pay attention to local informative areas and thus produces better results.

\subsubsection{The Feature Pyramid Fusion Module (FPFM)}

The feature pyramid fusion module is used to fuse shallow-layer feature with rich spatial information and deep-layer feature with rich semantic information. To investigate the benefits introduced by this module, we compare our DNA-Net with three variants, we increased the number of filters of all convolution layers of three variants to make their model sizes comparable for fair comparison.

\begin{itemize}
\item \textbf{DNA-Net w/o FPFM}: We replaced the feature pyramid fusion module in this variant and only used the output from the final layer as final result.
\item \textbf{DNA-Net w/o L345}: We removed the outputs of layer 3, 4, and 5 from FPFM in this variant to evaluate the contribution of features from middle and deep layers.
\item \textbf{DNA-Net w/o L45}:  We removed the outputs of layer 4, and 5 from FPFM in this variant to investigate the contribution of features from deep layers.
\item \textbf{DNA-Net w/o L5}:   We removed the outputs of layer 5 from FPFM in this variant to investigate the benefit introduced by the deepest layer of the network.
\end{itemize}

As shown in Table~\ref{Tab_FPFM}, \textit{DNA-Net w/o FPFM} suffers decreases of 0.83\%, 1.89\%, and a increase of 1.81 $\times 10^{-6}$ in terms of $IoU$, ${P}_{d}$, and ${F}_{a}$ on the NUDT-SIRST dataset. Similar results can also be observed on the NUAA-SIRST dataset. That is because, FPFM helps to achieve multi-layer features fusion. The representation from shallow layers and deep layers can be both extracted and fused to generate more robust feature maps as output.

When we gradually removed partial outputs of FPFM from bottom to the top layer, our network suffers decreases of 0.23\%, 0.84\%, and an increase of 3.01$\times 10^{-6}$ in terms of $IoU$, ${P}_{d}$, and ${F}_{a}$ for \textit{DNA-Net w/o L5}. Similar results can also be observed on \textit{DNA-Net w/o L45} and \textit{DNA-Net w/o L345}. That is because, NUDT-SIRST contains rich multi-target scenarios, more small size targets, and less visually salient targets. Our network can fully fuse low-level and high-level information and thus achieves better performance on NUDT-SIRST.

\subsection{Benefits of The Synthesized Dataset}\label{SOAT}

In this section, we evaluate the benefits of our synthesized dataset for real IRST tasks.
Specifically, we mixed real SIRST images (from the training set of NUAA-SIRST) and synthesized SIRST images (from the training set of NUDT-SIRST) with different ratios to train the networks and evaluated their performance on the real images (from the test set of NUAA-SIRST). As shown in Table~\ref{Tab_Potential_1}, with small ratio of real images, both DNA-Net and ACM can achieve comparable results to baseline results (trained on all real images). That is because, our synthesized dataset can well cover the main challenges for infrared small target detection (i.e., different SCR, clutter background, target shape, and target size). Consequently, the huge cost for collecting real SIRST images can be reduced.

\begin{figure}
\centering
\includegraphics[width=8.8cm]{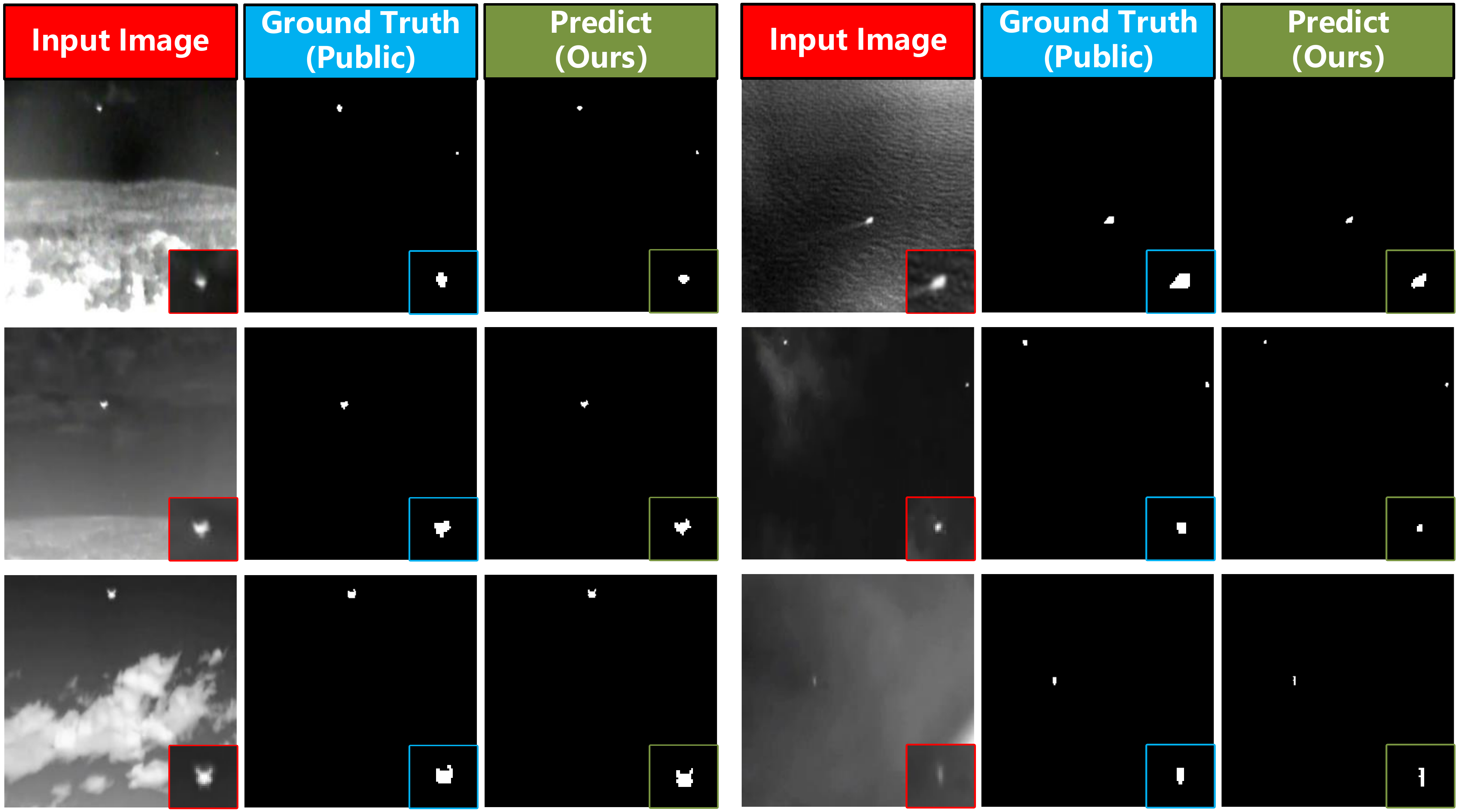}
\caption{Samples of the input images, public ground truth masks\cite{22-ACM} (manually labeled), and output of our DNA-Net trained on mixed dataset. Our method can even produce more precise segmentation result than manually labeled ground truth masks.}\label{Syn_adv}
\end{figure}

Moreover, we compared the output of our network trained on the mixed dataset with the manually labeled masks of NUAA-SIRST in Fig.~\ref{Syn_adv}. It can be observed that the outputs of our network have more reasonable shape segmentation than ground truth labels. That is because, the synthesized SIRST images have absolutely precise labels. The network can learn the essence of infrared small targets with sufficiently well labeled data and finally contribute to the improvement of real SIRST images. Our network can generate better visual performance than ground truth label.

\section{CONCLUSION}\label{SecConclusion}

\begin{table}[]\scriptsize
\centering
\renewcommand\arraystretch{1.1}
\caption{$IoU$($\times10^{2}$)$/$$P_{d}$($\times10^{2}$)$/$$F_{a}$($\times10^{6}$) values achieved by DNA-Net on real datasets. The DNA-Net is trained on mixed dataset with different real images ratios} \label{Tab_Potential_1}
\begin{tabular}{|p{1.7cm}<{\centering}|c|c|c|}
\hline
\multirow{2}{*}{\#Real image} & \multirow{2}{*}{\#Synthesized image} & \multicolumn{2}{c|}{Method}                   \\ \cline{3-4}
                       &                           & DNA-Net    & ACM                     \\ \hline
0\%  (0/791)     &   100\% (791/791)   &  66.84/95.43/28.59    & 60.43/91.25/20.75               \\ \hline
5.3\% (42/791)   &  94.7\%  (749/749)  &  70.44/96.19/15.40    & 63.94/92.87/27.37         \\ \hline
10.7\% (85/791)  &  89.3\%  (706/749)  &  74.58/97.43/7.263     & 66.69/93.67/16.95               \\ \hline
16.2\% (128/791) &  83.8\%  (663/749)  &  77.23/98.34/6.401    & 69.29/95.06/9.022             \\ \hline
100\% (213/213)  & -                   &  77.47/98.48/4.223    & 70.33/93.91/3.728             \\ \hline

\end{tabular}
\end{table}

In this paper, we propose a DNA-Net to achieve SIRST detection. Different from existing CNN-based SIRST detection methods, we explicitly handle the problem of small targets being lost in deep layers by designing a new tri-direction dense nested interactive module with a cascaded channel and spatial attention model. The intrinsic information of small targets can be incorporated and fully exploited by repeated fusion and enhancement. Moreover, we develop an open SIRST dataset to evaluate the performance of infrared small target detection with respect to challenging scenes. We also reorganized a set of evaluation metrics. Experiments on both our dataset and the public dataset have shown the superiority of our method over the state-of-the-art methods.

\bibliographystyle{IEEEtran}
\bibliography{DNANet}

\begin{IEEEbiography}[{\includegraphics[width=1in,height=1.25in,clip,keepaspectratio]{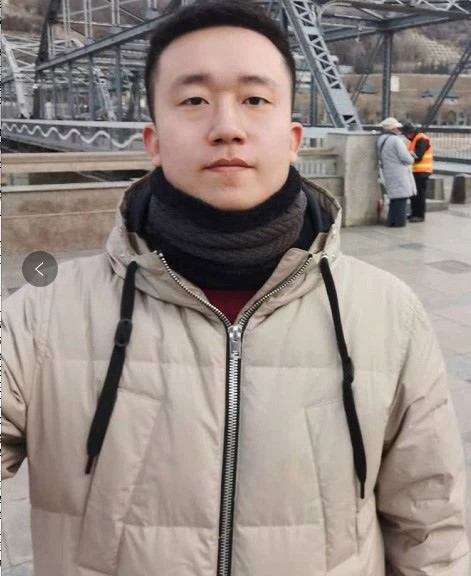}}]{Boyang Li} received the B.E. degree in Mechanical Design manufacture and Automation from the Tianjin University, China, in 2017 and M.S. degree in biomedical engineering from National Innovation Institute of Defense Technology, Academy of Military Sciences, Beijing, China, in 2020. He is currently working toward the PhD degree in information and communication engineering from National University of Defense Technology (NUDT), Changsha, China. His research interests include infrared small target detection, weakly supervised semantic segmentation and deep learning.
\end{IEEEbiography}

\begin{IEEEbiography}[{\includegraphics[width=1in,height=1.25in,clip,keepaspectratio]{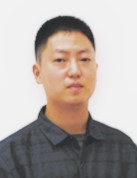}}]{Chao Xiao} received the BE degree in the communication engineering and the ME degree in information and communication engineering from the National University of Defense Technology (NUDT), Changsha, China in 2016 and 2018, respectively. He is currently working toward the Ph.D. degree with the College of Electronic Science in NUDT, Changsha, China. His research interests include deep learning, small object detection and multiple object tracking.
\end{IEEEbiography}

\begin{IEEEbiography}[{\includegraphics[width=1in,height=1.25in,clip,keepaspectratio]{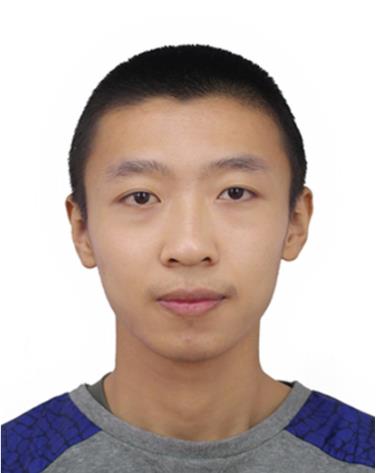}}]{Longguang Wang} received the B.E. degree in electrical engineering from Shandong University (SDU), Jinan, China, in 2015, and the M.E. degree in information and communication engineering from National University of Defense Technology (NUDT), Changsha, China, in 2017. He is currently pursuing the Ph.D. degree with the College of Electronic Science and Technology, NUDT. His research interests include low-level vision and deep learning.
\end{IEEEbiography}

\begin{IEEEbiography}[{\includegraphics[width=1in,height=1.25in,clip,keepaspectratio]{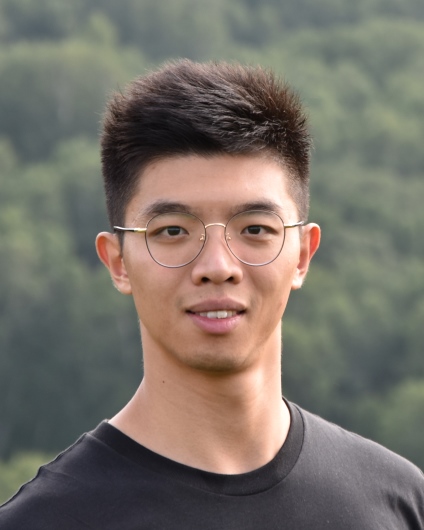}}]{Yingqian Wang} received the B.E. degree in electrical engineering from Shandong University (SDU), Jinan, China, in 2016, and the M.E. degree in information and communication engineering from National University of Defense Technology (NUDT), Changsha, China, in 2018. He is currently pursuing the Ph.D. degree with the College of Electronic Science and Technology, NUDT. His research interests focus on low-level vision, particularly on light field imaging and image super-resolution.
\end{IEEEbiography}

\begin{IEEEbiography}[{\includegraphics[width=1in,height=1.25in,clip,keepaspectratio]{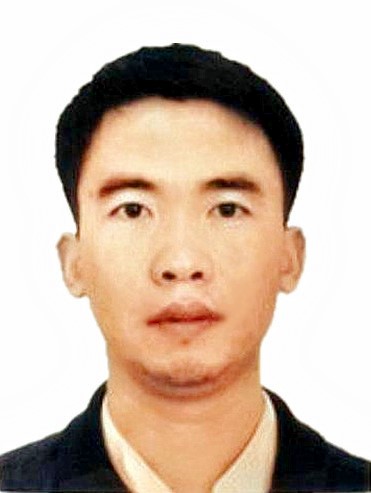}}]{Zaiping Lin} received the B.Eng. and Ph.D. degrees from the National University of Defense Technology (NUDT) in 2007 and 2012, respectively. He is currently an Associate Professor with the College of Electronic Science and Technology, NUDT. His current research interests include infrared image processing and signal processing.
\end{IEEEbiography}

\begin{IEEEbiography}[{\includegraphics[width=1in,height=1.25in,clip,keepaspectratio]{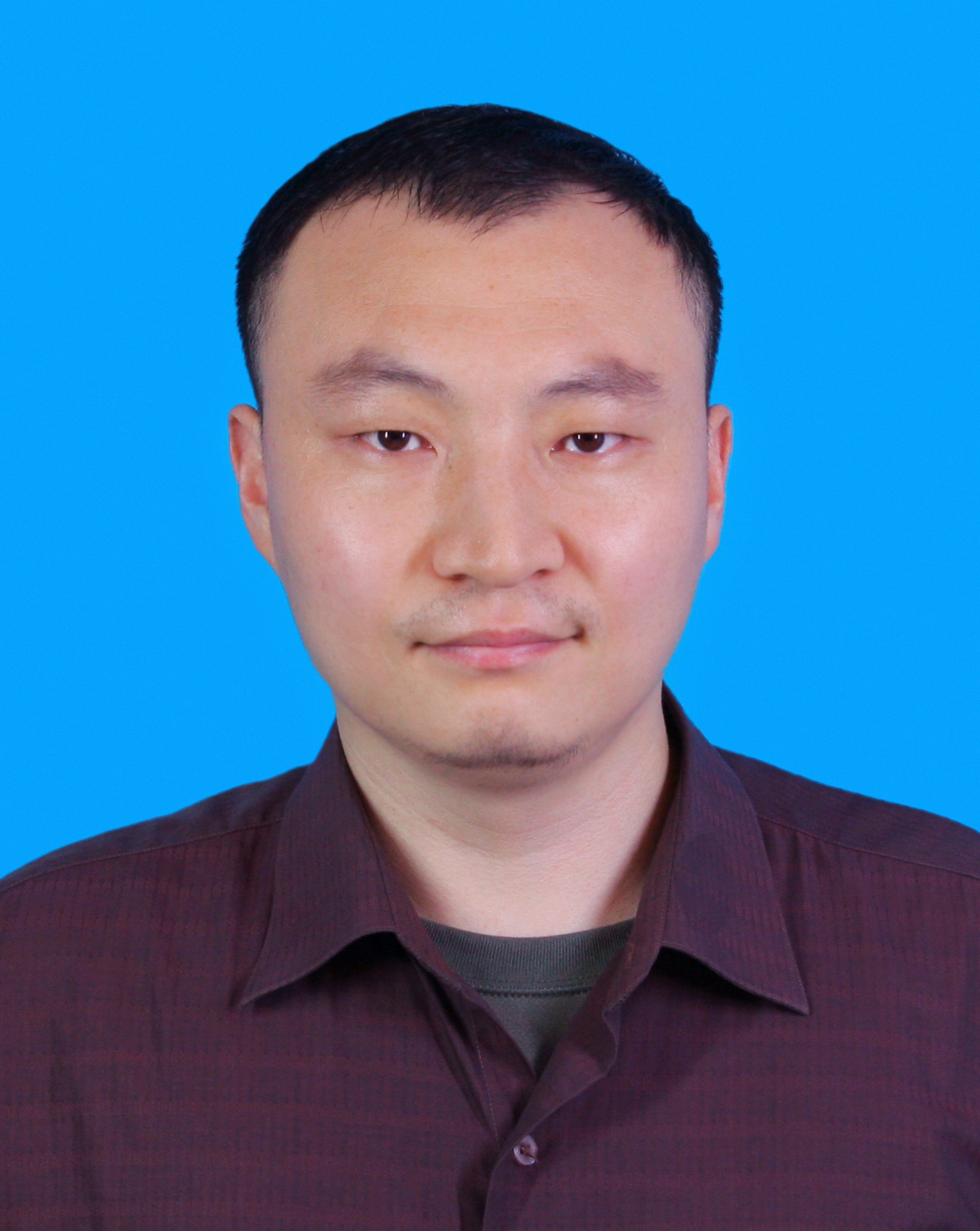}}]{Miao Li} received the M.E. and Ph.D. degrees from the National University of Defense Technology (NUDT) in 2012 and 2017, respectively. He is currently an Associate Professor with the College of Electronic Science and Technology, NUDT. His current research interests include infrared dim and small target detection.
\end{IEEEbiography}

\begin{IEEEbiography}[{\includegraphics[width=1in,height=1.25in,clip,keepaspectratio]{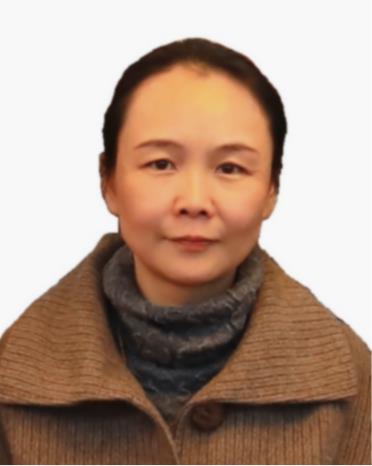}}]{Wei An} received the Ph.D. degree from the National University of Defense Technology (NUDT), Changsha, China, in 1999. She was a Senior Visiting Scholar with the University of Southampton, Southampton, U.K., in 2016. She is currently a Professor with the College of Electronic Science and Technology, NUDT. She has authored or co-authored over 100 journal and conference publications. Her current research interests include signal processing and image processing.
\end{IEEEbiography}

\begin{IEEEbiography}[{\includegraphics[width=1in,height=1.25in,clip,keepaspectratio]{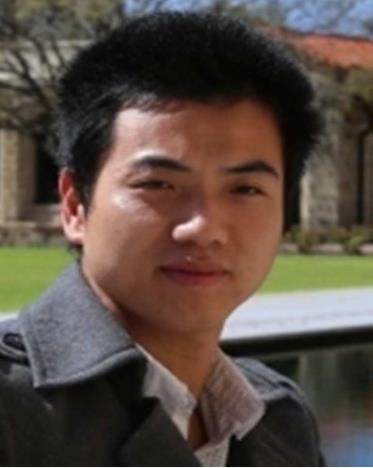}}]{Yulan Guo} received the B.E. and Ph.D. degrees from National University of Defense Technology (NUDT) in 2008 and 2015, respectively. He has authored over 100 articles at highly referred journals and conferences. His current research interests focus on 3D vision, particularly on 3D feature learning, 3D modeling, 3D object recognition, and scene understanding. He served as an associate editor for IEEE Transactions on Image Processing, IET Computer Vision, IET Image Processing, and Computers \& Graphics. He also served as an area chair for CVPR 2021, ICCV 2021, and ACM Multimedia 2021. He organized several tutorials, workshops, and challenges in prestigious conferences, such as CVPR 2016, CVPR 2019, ICCV 2021, 3DV 2021, CVPR 2022, ICPR 2022, and ECCV 2022. He is a Senior Member of IEEE and ACM.
\end{IEEEbiography}

\end{document}